\pdfoutput=1
\documentclass[11pt,a4paper]{article}
\usepackage[final]{emnlp2021}
\usepackage{times}
\usepackage{latexsym}

\usepackage{amsmath,mathtools,amsfonts}
\usepackage{amssymb}
\usepackage{booktabs}
\usepackage{bbm}
\usepackage{multirow}
\usepackage{rotating}
\usepackage[outline]{contour}
\usepackage{array}

\usepackage{caption} 
\usepackage{subcaption}

\newcommand{\mypara}[1]{\noindent\textbf{#1}}

\usepackage[T1]{fontenc}
\usepackage[utf8]{inputenc}
\usepackage{microtype}
\usepackage{floatrow}
\usepackage{pifont}
\newcommand{\cmark}{\ding{51}}%
\newcommand{\xmark}{\ding{55}}%

\definecolor{darkgreen}{rgb}{0.13, 0.55, 0.13}

\newcommand{\M}{$\mathcal{M}$}
\newcommand{\F}{$\mathcal{F}$}
\newcommand{\FW}{$\mathcal{F}_{\text{W}}$}
\newcommand{\FC}{$\mathcal{F}_{\text{C}}$}
\newcommand{\T}{$\tau$}
\newcommand{\TM}{$\tau_1$}
\newcommand{\TF}{$\tau_2$}
\newcommand{\TFBest}{${\tau_2}^*$}
\newcommand{\TT}{\tau}
\newcommand{\R}{$\mathcal{R}$}
\newcommand{\MT}{$\mathcal{M}\langle\tau\rangle$}
\newcommand{\MTM}{$\mathcal{M}\langle\tau_1\rangle$}
\newcommand{\MTT}{\mathcal{M}\langle\tau\rangle}
\newcommand{\FMT}{$\mathcal{F}_{\mathcal{M}\langle\tau\rangle}$}
\newcommand{\FMTM}{$\mathcal{F}_{\mathcal{M}\langle\tau_1\rangle}$}
\newcommand{\FMTval}[1]{$\mathcal{F}_{\mathcal{M}\langle{#1}\rangle}$}
\newcommand{\Mval}[1]{$\mathcal{M}\langle{#1}\rangle$}
\newcommand{\FM}{$\overline{{\mathcal{{F}}_{\mathcal{M}}}}$}

\usepackage{todonotes}
\usepackage{enumitem}
\setlist[itemize]{align=parleft,left=0pt..1em}

\title{\emph{Will this Question be Answered?} Question Filtering via Answer Model Distillation for Efficient Question Answering}

\author{Siddhant Garg \and Alessandro Moschitti\\
Amazon Alexa AI\\
 \texttt{\{sidgarg,amosch\}@amazon.com} \\
}

\begin{document}
\maketitle
\begin{abstract}
In this paper we propose a novel approach towards improving the efficiency of Question Answering (QA) systems by filtering out questions that will not be answered by them. This is based on an interesting new finding: the answer confidence scores of state-of-the-art QA systems can be approximated well by models solely using the input question text. This enables preemptive filtering of questions that are not answered by the system due to their answer confidence scores being lower than the system threshold. Specifically, we learn Transformer-based question models by distilling Transformer-based answering models. Our experiments on three popular QA datasets and one industrial QA benchmark demonstrate the ability of our question models to approximate the Precision/Recall curves of the target QA system well. These question models, when used as filters, can effectively trade off lower computation cost of QA systems for lower Recall, e.g., reducing computation by ${\sim}60\%$, while only losing ${\sim}3{-}4\%$ of Recall.
\end{abstract}

\section{Introduction}
Question Answering (QA) technology is at the core of several commercial applications, e.g., virtual assistants such as Alexa, Google Home and Siri, serving millions of users.
Optimizing the efficiency of such systems is vital to reduce their operational costs. Recently, there has been a large body of research devoted towards reducing the compute complexity of retrieval~\cite{10.1145/3289600.3290986,tan-etal-2019-efficient} and transformer-based QA models~\cite{DBLP:journals/corr/abs-1910-01108,soldaini-moschitti-2020-cascade}.

An alternate solution for improving QA system efficiency aims to discard questions that will most probably be incorrectly answered by the system, using automatic classifiers. For example, \citet{fader-etal-2013-paraphrase,faruqui-das-2018-identifying} aim to capture the grammaticality and well-formedness of questions. 
However, these methods do not take the specific answering ability of the target system into account. In practice, QA systems typically do not answer a significant portion of user questions since their answer scores could be lower than a confidence threshold~\cite{kamath-etal-2020-selective}, tuned by the system to achieve the required Precision. 
For example, QA systems for medical domains exhibit a high Precision since providing incorrect/imprecise answers can have critical consequences for the end user. Based on the above rationale, discarding questions that will not be answered by the QA system presents a remarkable cost-saving opportunity. However, applying this idea may appear unrealistic from the outset since the QA system must first be executed to generate the answer confidence score.

In this paper, we take a new perspective on improving QA system efficiency by preemptively filtering out questions that will \emph{not} be answered by the system, by means of a question filtering model. This is based on our interesting finding that the answer confidence score of a QA system can be well approximated solely using the question text.

Our empirical study is supported by several observations and intuitions. First, the final answer confidence score from a QA system (irrespective of its complex pipeline) is often generated by a Transformer-based model. This is because Transformer-based models are used for answer extraction in most research areas in QA with unstructured text, e.g., Machine Reading(MR)~\cite{rajpurkar-etal-2016-squad}, and Answer Sentence Selection(AS2)~\cite{Garg_Vu_Moschitti_2020}.
Second, more linguistically complex questions have a lower probability to be answered. Language complexity correlates with syntactic, semantic and lexical properties, which have been shown to be well captured by pre-trained language models (LMs) \cite{jawahar-etal-2019-bert}. Thus, the final answer extractor will be affected by said complexity, suggesting that we can predict which questions are likely to be unanswered just using their surface forms. 

Third, pre-training transformer-based LMs on huge amounts of web data enables them to implicitly capture the frequency/popularity of general phrases\footnote{Intended as general sequence of words, not necessarily specific to a grammatical theory, which LMs can capture well.}, among which entities and concepts play a crucial role for answerability of questions. Thus, the contextual embedding of a question from a transformer LM is, to some extent, aware of the popularity of entities and concepts in the question, which impacts the retrieval quality of \emph{good} answer candidates. This means that a portion of the retrieval complexity of a QA system can also be estimated just using the question.
Most importantly, we only try to estimate the answer score from a QA system and not whether the answer provided by the system for a question is correct or incorrect (the latter being a much more difficult task).

Following the above intuitions, we distill the knowledge of QA models, using them as teachers, into Transformer-based models (students) that only operate on the question. Once trained, the student question model can be used to preemptively filter out questions whose answer score will not clear the system threshold, translating to a proportional reduction in the runtime cost of the system. More specifically, we propose two loss objectives for training two variants of this question filter: one with a regression head and one with a classification head. The former attempts to directly predict the continuous score provided by the QA system.  The latter aims at learning to predict if a question will generate a score ${>}\tau$, which is the answer confidence threshold the QA system was tuned to.

We perform empirical evaluation for (i) showing the ability of our question models to estimate the QA system score; and (ii) testing the cost savings produced by our question filters, trading off with a drop in Recall.
We test our models on two QA tasks with unstructured text, MR and AS2, using (a) three academic datasets: WikiQA, ASNQ, and SQuAD 1.1; (b) a large scale industrial benchmark, and (c) a variety of different transformer architectures such as BERT, RoBERTa and ELECTRA. 
Specifically for (i), we compare the Precision(Pr)/Recall(Re) curves of the original and the new QA system, where the latter uses the question model score to trade-off Precision for Recall. For (ii), we show the cost savings produced by our question filters, when operating the original QA system at different Precision values. The results show that:
(i) The Pr/Re curves of the question models are close to those of the original system, suggesting that they can estimate the system scores well; and 
(ii) our question models can preemptively filter out $21.9{-}45.8\%$ questions while only incurring a drop in Recall of $3.2{-}4.9\%$. \footnote{Code will be released soon at \url{https://github.com/alexa/wqa-question-filtering} } 

\section{Related Work}
\label{relwork}
\mypara{Question Answering} Prior efforts on QA have been broadly categorized into two fronts: tackling MR, and AS2. For the former, recently pre-trained transformer models~\cite{devlin-etal-2019-bert,liu-19-roberta,Lan2020ALBERT,clark2020electra}, etc. have achieved SOTA performance, sometimes even exceeding the human performance. Progress on this front has also seen the development of large-scale QA datasets like SQuAD~\cite{rajpurkar-etal-2016-squad}, HotpotQA~\cite{yang2018hotpotqa}, NQ~\cite{Kwiatkowski_NQ}, etc.~with increasingly challenging types of questions. For the task of AS2, initial efforts embedded the question and candidates using CNNs~\cite{Severyn:2015:LRS:2766462.2767738}, weight aligned networks~\cite{shen-etal-2017-inter,tran-etal-2018-context,DBLP:journals/corr/abs-1806-00778} and compare-aggregate architectures~\cite{DBLP:journals/corr/WangJ16b,Bian:2017:CMD:3132847.3133089,DBLP:journals/corr/abs-1905-12897}. Recent progress has stemmed from the application of transformer models for performing AS2~\cite{Garg_Vu_Moschitti_2020,han-etal-2021-modeling,10.1007/978-3-030-72113-8_20}. 
On the data front, small datasets like TrecQA~\cite{wang-etal-2007-jeopardy} and WikiQA~\cite{yang-etal-2015-wikiqa} have been supplemented with datasets such as ASNQ~\cite{Garg_Vu_Moschitti_2020} having several million QA pairs. 

Open Domain QA (ODQA)~\cite{chen2017reading,chen-yih-2020-open} systems involve a combination of a retriever and a reader~\cite{semnani2020revisiting} trained independently~\cite{10.1145/3077136.3080721} or jointly~\cite{yang-etal-2019-end-end}. 
Efforts in ODQA transitioned from using knowledge bases for answering questions to using external text sources and web articles~\cite{10.1145/2911451.2911536,sun-etal-2018-open,xiong-etal-2019-improving,10.1145/3331184.3331252}. 
Numerous research works have proposed different techniques for improving the performance on ODQA~\cite{DBLP:journals/corr/abs-1911-03868,DBLP:journals/corr/abs-1911-10470,10.1145/3331184.3331190,qi2019answering}. 

\mypara{Filtering Ill-formed Questions}
Evaluating well-formedness and intelligibility of queries has been a popular research topic for QA systems. \citeauthor{faruqui-das-2018-identifying} annotate the Paralex dataset~\cite{fader-etal-2013-paraphrase} on the well-formedness of the questions. The majority of research efforts have been aimed at reformulating user queries to elicit the best possible answer from the QA system~\cite{10.1145/2631775.2631809,DBLP:journals/corr/BuckBCGHGW17,DBLP:journals/corr/abs-1911-09247}. 
A complementary line of work uses hate speech detection techniques~\cite{9297560} to filter questions that incite hate on the basis of race, religion, etc.

\mypara{Answer Verification}
QA systems sometimes use an answer validation component in addition to the system threshold, which analyzes the answer produced by the system and decides whether to answer or abstain. These systems often use external entity knowledge~\cite{magnini-etal-2002-right,ko-etal-2007-probabilistic,GondekLallyEtAl12ibmjrd} for basing their decision to verify the correctness of the answer. Recently ~\citet{wang-etal-2020-answer-better} propose to add a new MR model to reflect on the predictions of the original MR model to decide whether the produced answer is correct or not. Other efforts~\cite{DBLP:journals/corr/abs-1904-04792,kamath-etal-2020-selective,Jia2020KnowWT,zhang-etal-2021-knowing} have trained calibrators for verifying if the question should be answered or not. All these works are fundamentally different from our question filtering approach since they operate jointly on the question and generated answer, thereby requiring the entire computation to be performed by the QA system before making a decision. Our work operates only on the question text to preemptively decide whether to filter it or not. Thus the primary goal of these existing works is to improve the precision of the answering model by not answering when not confident, while our work aims to improve efficiency of the QA system and save runtime compute cost.

\mypara{Query Performance Prediction}
Pre-retrieval query difficulty prediction has been previously explored in Information Retrieval~\cite{10.1145/1835449.1835683}. Previous works~\cite{10.1007/978-3-540-30213-1_5,Mothe05linguisticfeatures,10.1007/978-3-540-78646-7_80,10.1007/978-3-540-78646-7_8,10.1145/1842890.1842906} target $p(a|q,f)$, ground truth probability of an answer $a$ to be correct, given a question $q$ and a feature set $f$ in input using simple linguistic (e.g., parse trees, polysemy value) and statistical (e.g., query term statistics, PMI) methods; while we target the QA-system score $s(a|q,f)$, the probability of an answer to be correct \emph{as estimated} by the system. This task is more semantically driven than syntactically, and enables the use of large amounts of training data without human labels of answer correctness. 

\mypara{Efficient QA}
Several works on improving the efficiency of the retrieval involve using a cascade of re-rankers to quickly identify good documents~\cite{10.1145/2009916.2009934, 10.1145/2911451.2911515, 10.1145/3289600.3290986}, non-metric matching functions for efficient search~\cite{tan-etal-2019-efficient}, etc.
Towards reducing compute of the answer model, the following techniques have been explored: multi-stage ranking using progressively larger models~\cite{10.1145/3397271.3401266}, using intermediate representations for early elimination of negative candidates~\cite{soldaini-moschitti-2020-cascade,xin-etal-2020-deebert}, combining separate encoding of question and answer with shallow DNNs~\cite{Chen_2020}, and the most popular being knowledge distillation~\cite{HinVin15Distilling} to train smaller transformer models with low inference latencies (DistilBERT~\cite{DBLP:journals/corr/abs-1910-01108}, TinyBERT~\cite{jiao2020tinybert}, MobileBERT~\cite{sun-etal-2020-mobilebert}, etc.)

\section{Preliminaries and Problem Setting}
\label{prob}

\begin{figure}[t]
    \centering
    \includegraphics[width=0.9\linewidth]{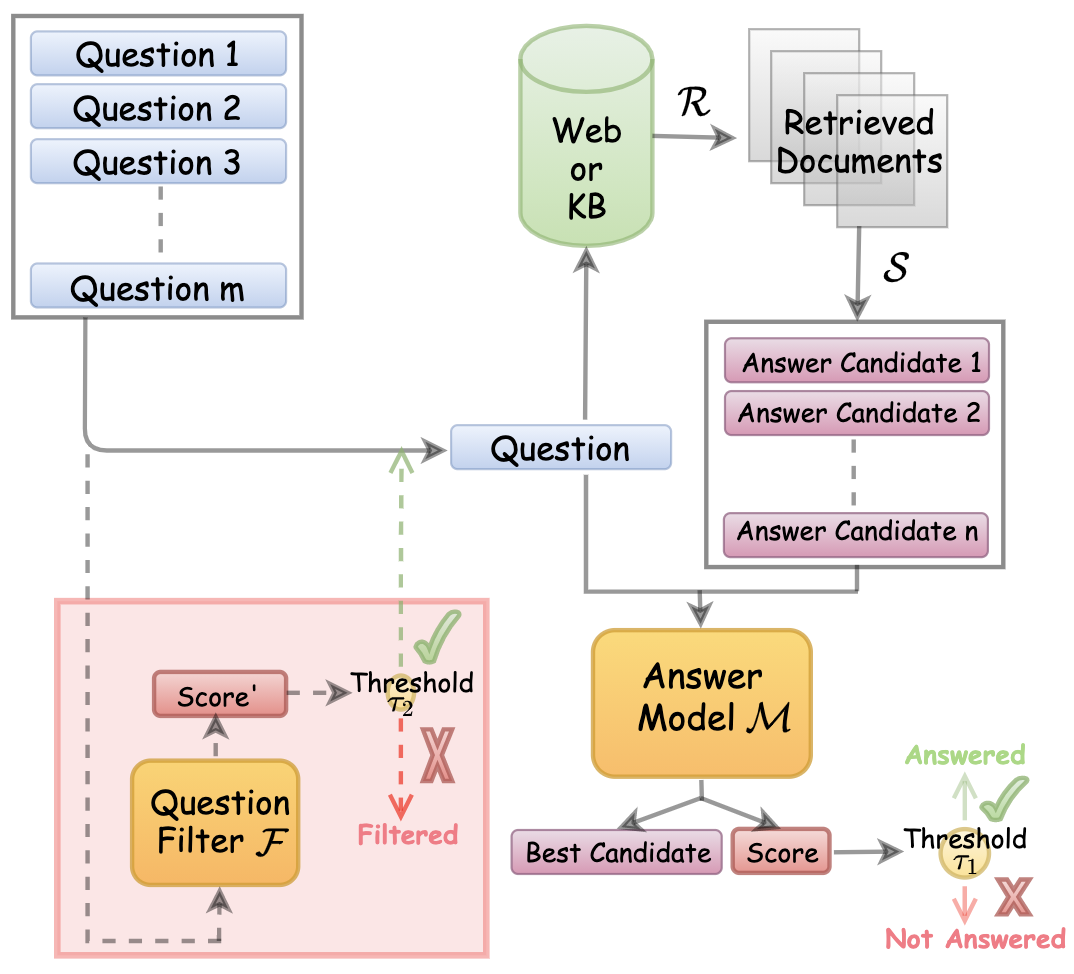}
    \vspace{-0.25cm}
    \caption{A real-world QA system having a retrieval ($\mathcal{R}$), candidate extraction ($\mathcal{S}$) and answering component ({\M}). Our proposed question filter (highlighted by the red box) preemptively removes the questions which will fail the threshold {\TM} of {\M}.}
    \label{fig:qa_system}
\end{figure}

We first provide details of QA systems and explain the cost-saving opportunity space when they operate at a given Precision (or answer score threshold).

\subsection{QA Systems for Unstructured Text}
We consider QA systems based on unstructured text, a simple design for which works as follows (as depicted in Fig.~\ref{fig:qa_system}): given a user question $q$, a search engine, $\mathcal{R}$, first retrieves a set of documents (e.g., from a web index). A text splitter, $\mathcal{S}$, applied to the documents, produces a set of passages/sentences, which are then input to an answering model $\mathcal{M}$. The latter produces the final answer. There are two main research areas studying the design of $\mathcal{M}$:

\mypara{Machine Reading (MR):} $\mathcal{S}$ extracts passages $\{p_1,{\dots},p_m\}$ from the retrieved documents. $\mathcal{M}$ is a reading comprehension head, which uses $(q,\{p_1,{\dots},p_m\})$ to predict start and end position (span) for the best answer based on these passages.

\mypara{Answer Sentence Selection (AS2):} $\mathcal{S}$ splits the retrieved documents into a set $\{s_1,{\dots},s_m\}$ of individual sentences. $\mathcal{M}$ performs sentence re-ranking over $(q,\{s_1,{\dots},s_m\})$, where the top ranked candidate is provided as the final answer. $\mathcal{M}$ is typically learned as a binary classifier applied to QA pairs, labelled as being correct or incorrect. AS2 models can handle scaling to large collection of sentences (more documents and candidates) more easily than MR models (the latter process passages in entirety before answering while the former can break down passages into candidates and evaluate them in parallel) thereby having lower latency at inference time.
\vspace{-0.25cm}

\subsection{Precision/Recall Tradeoff}
\vspace{-0.01cm}
{\M} provides the best answer (irrespective of the answer modeling being MR or AS2) for a given question $q$ along with a prediction score $\sigma$ (DNNs typically produce a normalized probability), which is termed \textit{MaxProb} in several works~\cite{hendrycks17baseline,kamath-etal-2020-selective}. 
The most popular technique to tune the Pr/Re tradeoff is to set a threshold, {\T}, on $\sigma$. 
This means that the system provides an answer for $q$ only if $\sigma{>}$ {\T}. 
Henceforth, we denote {\M} operating at a threshold {\T} by {\MT}. 
While not calibrated perfectly (as shown by~\citeauthor{kamath-etal-2020-selective}), the predictions of QA models are supposed to be aligned with the ground truth such that questions that are \emph{correctly} answered are more likely to receive higher $\sigma$ than those that are \emph{incorrectly} answered. 
This is an effect of the binary cross-entropy loss, ${\cal L}_{\text{CE}}$, typically used for training {\M}\footnote{For MR, the sum of two cross entropy loss values is used: one for the start and one for the end of the answer. This sum is not exactly the probability of having a correct/incorrect answer, but correlates well with the probability of correctness.}.
For example, Fig.~\ref{fig:calibration_plots} plots Pr/Re on varying threshold {\T} of popular MR and AS2 systems, both built using transformer models (SQuAD: BERT-Base {\M}, ASNQ: RoBERTa-Base {\M}, details in Section~\ref{subsec:models}). The results show that increasing {\T} achieves a higher Pr trading it off for lower Re.

\subsection{Question Filtering Opportunity Space}
\begin{figure}[t]
\begin{subfigure}[t]{\linewidth}
    \begin{subfigure}[t]{0.49\textwidth}
        \includegraphics[width=\textwidth]{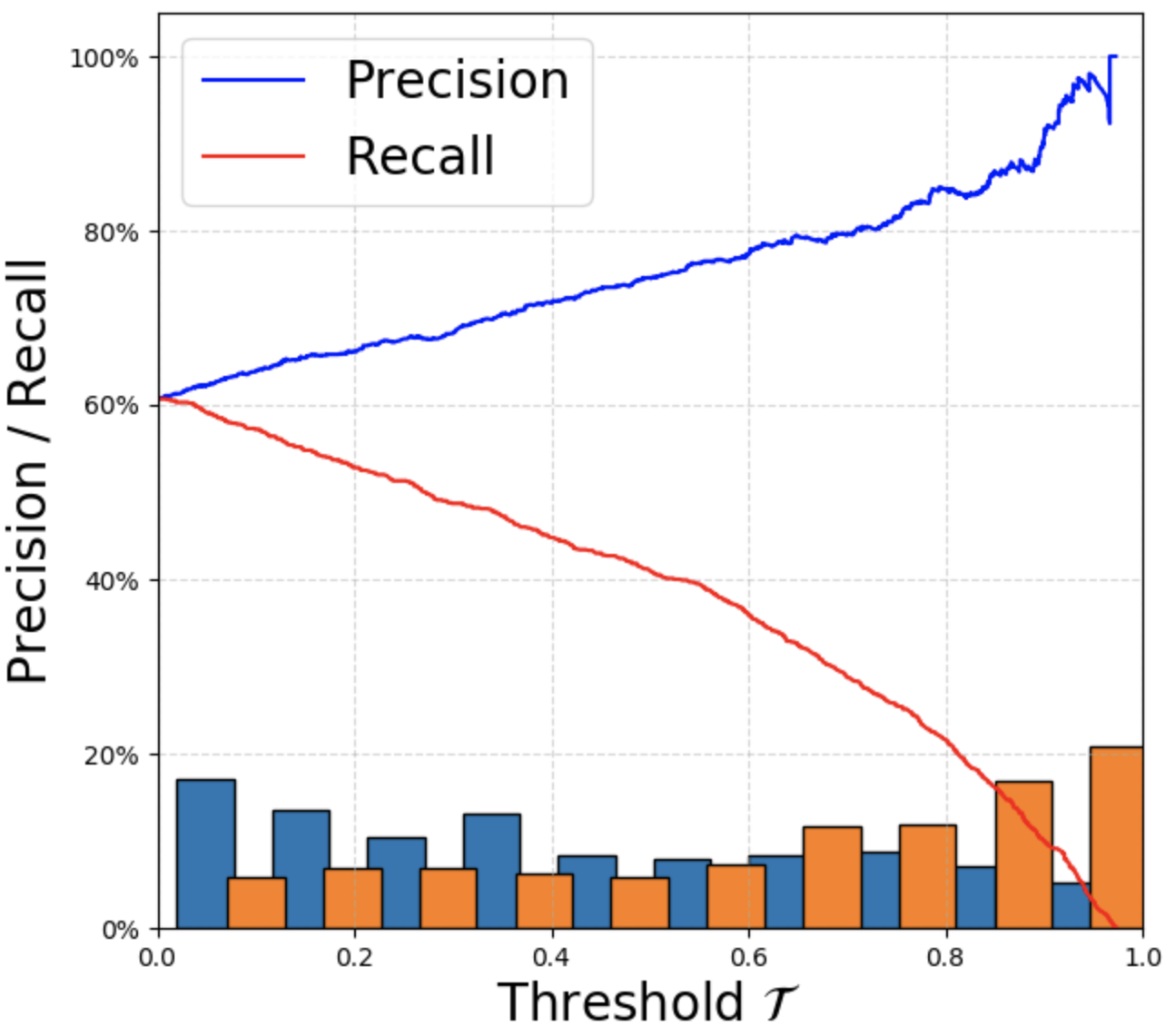}
        \caption{ASNQ}
    \end{subfigure}
    \begin{subfigure}[t]{0.49\textwidth}
        \includegraphics[width=\textwidth]{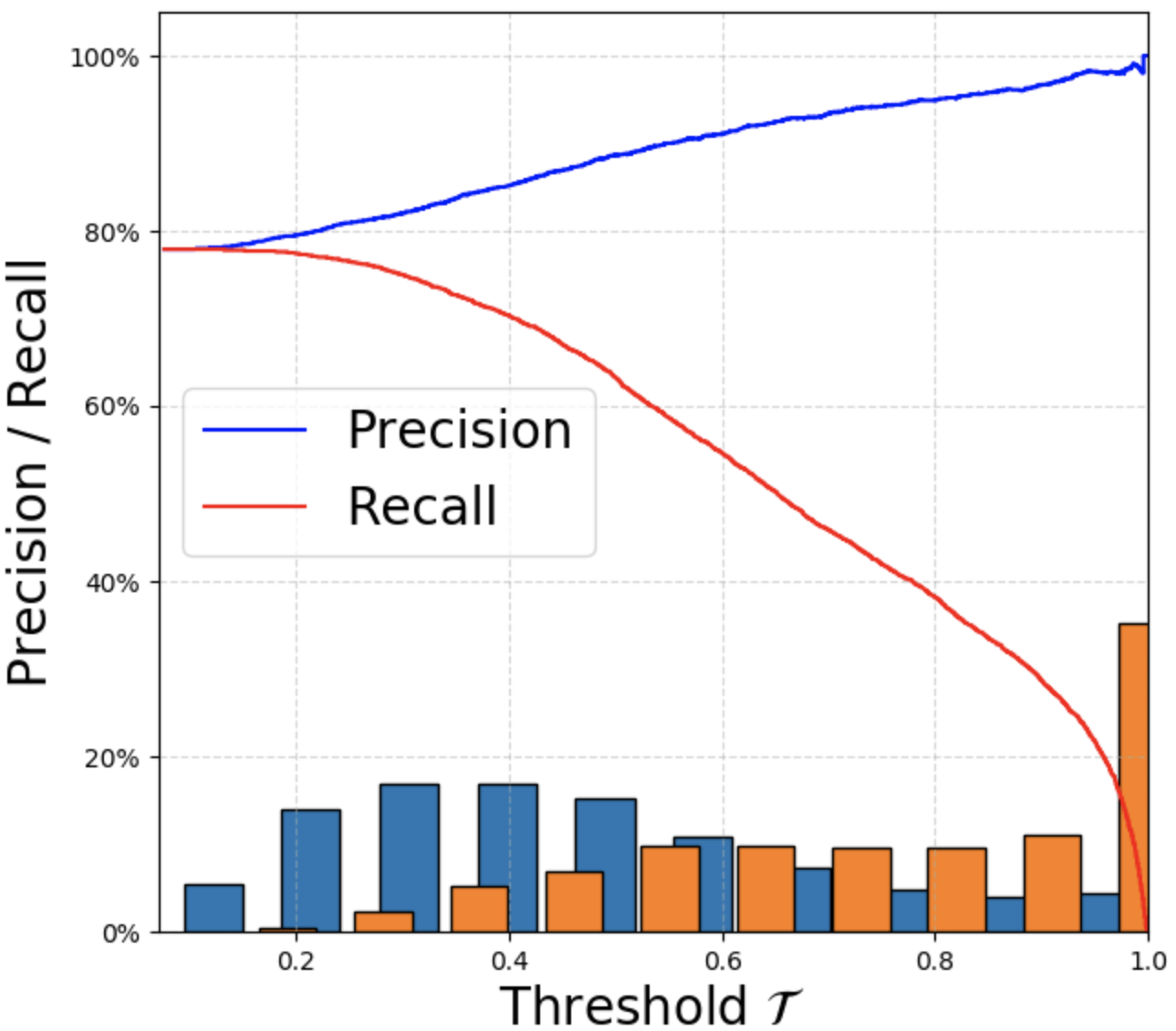}
        \caption{SQuAD 1.1}
    \end{subfigure}
\end{subfigure}
    \vspace{-0.25cm}
    \caption{Change in Pr/Re on varying threshold {\T} for {\M}. Additionally, we plot fraction of \textcolor{orange}{correctly}/ \textcolor{blue}{incorrectly} answered questions with the score $\sigma$ of {\M} in ranges $[0,0.1],{\dots},[0.9,1]$ on the x-axis. Frequency of \textcolor{orange}{correct} answers increases towards the right (as $\sigma{\rightarrow} 1$).}
    \label{fig:calibration_plots}
\end{figure}

Real-world QA systems are always associated with a target Precision, which is typically rather high to meet the customer quality requirements\footnote{Even if the Recall were very low this system can still be very useful for serving a portion of customer requests, as a part of a committee of specialized QA systems each answering specific user requests.}
using the threshold {\T}.
This means that systems will not provide an answer for a large fraction of questions ($q$'s for which $\sigma \le$ {\T}). For example from Fig.~\ref{fig:calibration_plots}(b), to obtain a Precision of $90\%$ for SQuAD 1.1, we need to set $\tau{=}0.55$, resulting in not answering $35.2\%$ of all questions. Similarly, to achieve the same Precision on ASNQ (Fig.~\ref{fig:calibration_plots}(a)), we need to set $\tau{=}0.89$, resulting in not answering $88.8\%$ of all questions. 

It is important to note that the QA system still performs the entire computation: {\R}${\rightarrow}\mathcal{S}{\rightarrow}${\M} on all questions (even the unanswered ones), to decide whether to answer or not. Thus, filtering these questions before executing the system can save the cost of running redundant computation, e.g., 35.2\% or 88.8\% of the cost of the two systems above (assuming the required Precision value is $90\%$). 
In the next section, we show how we build models that can produce a reliable prediction of the QA system score, only using the question text.

\section{Modelling QA System Scores using Input Questions}
\label{model}

We propose to use a distillation approach to learn a model operating on questions that can predict the confidence score of the QA system, within a certain error bound. We denote the QA system with $\Omega(\mathcal{R},\mathcal{S},${\MT}$)$, and  the question model by {\F} (as we will use it to filter out questions preemptively). 
Intuitively, {\F} aims at learning how confident the answer model {\M} is on answering a particular question when presented with a set of candidate answers from a retrieval system $(\mathcal{R},\mathcal{S})$. 

\begin{figure}[t]
    \centering
    \includegraphics[width=.9\linewidth]{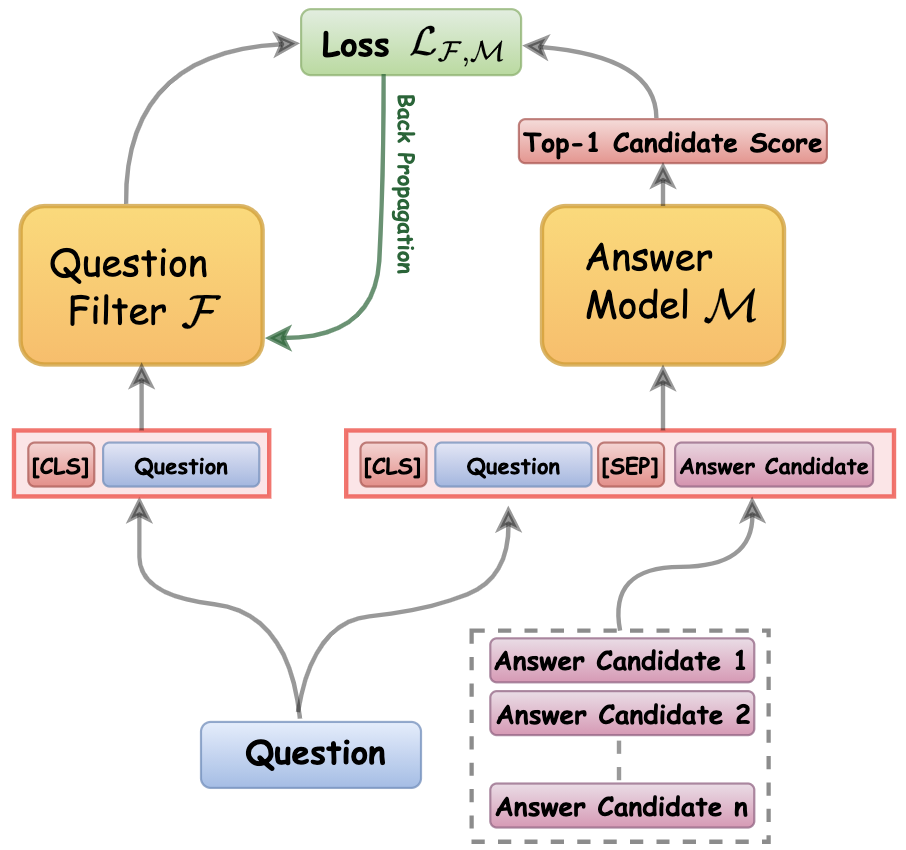}
    \vspace{-0.125cm}
    \caption{Distilling QA model {\M} to train the question filter {\FM} with a regression head using the MSE $\mathcal{L}_{\mathcal{F},\mathcal{M}}$ }
    \label{fig:distillation}
\end{figure}

For a question $q$, we indicate the set of answer candidate sentences/passages by $\bar{s}=\{s_1,{\dots},s_m\}$. The output from {\M} for $q$ and $\bar{s}$ corresponds to the score for the best candidate/span:
$\mathcal{M}(q,\bar{s}) = \max_{s \in \bar{s}} \mathcal{M}(q,s)$. 
We train a filter {\FM} for {\M} using a regression head to directly predict the score of {\M} irrespective of the threshold {\T} using the loss: 
\begin{align}
\label{eqn:regression_loss}
\mathcal{L}_{\mathcal{F},\mathcal{M}} (q,\bar{s}) {=} {\cal L}_{\text{MSE}}\big(\mathcal{F}(q),\mathcal{M}(q,\bar{s})\big),
\end{align}
\noindent where ${\cal L}_{\text{MSE}}$ is the mean square error loss. 
Fig.~\ref{fig:distillation} diagrammatically shows the training process of {\FM}.

Additionally, as {\M} typically operates with a threshold {\T}, we train a filter {\FMT} corresponding to a specific {\T}, i.e., {\MT}, using the following loss:
\begin{align}
\label{eqn:classifier_loss}
\hspace{-.7em} \mathcal{L}_{\mathcal{F},\MTT} (q,\bar{s}) {=} {\cal L}_{\text{CE}} \Big(\mathcal{F}(q),\mathbbm{1}\big(\mathcal{M}(q,\bar{s}) {>} \TT \big)\Big),
\end{align}

\noindent where ${\cal L}_{\text{CE}}$ is the cross entropy loss and $\mathbbm{1}$ denotes the binary indicator function. 

The novelty of our proposed approach from standard distillation techniques \cite{HinVin15Distilling,DBLP:journals/corr/abs-1910-01108,jiao2020tinybert} stems from the fact that, unlike the standard setting, in our case the teacher {\M} and the student {\F} operate on different inputs: {\F} only on the questions while {\M} on question-answer pairs. This makes our task much more challenging as {\F} needs to approximate the probability of {\M} fitting \emph{all} answer candidates for a question. Since our {\F} does not predict if an answer provided by {\M} is correct/incorrect, we don't require labels of the QA dataset for training {\F} ({\F}'s output only depends on predictions of {\M}). This enables large scale training of {\F} without any human supervision using the predictions of a system $\Omega(\mathcal{R},\mathcal{S},${\MT}$)$ and a large number of questions. 

To use the trained {\F} for preemptively filtering out questions, we use a threshold on the score of {\F}. Henceforth, we refer to the threshold of {\M} by {\TM} and that of {\F} by {\TF}. Any question $q$ for which {\F}$(q) {\le} \tau_2$, gets filtered out. Using the question filter, we define the new QA system as $\widehat\Omega(\mathcal{F}{\langle\tau_2\rangle},\mathcal{R},\mathcal{S},{\mathcal{M}}{\langle\tau_1\rangle})$ where the filter {\F} can be trained using Eq.~\ref{eqn:regression_loss} or \ref{eqn:classifier_loss} ($\overline{\mathcal{F}_{\mathcal{M}}}$ or $\mathcal{F}_{\mathcal{M}\langle \tau_1 \rangle}$).

\section{Experiments}
\label{exp}
First, we compare how well our models {\F} can approximate the answer score of {\M}.
Then we optimize $\tau_1$ and $\tau_2$ on the dev.~set to precisely estimate the cost savings that we can obtain with the application of our approach to questions filtering. We also compare it with different baseline models for $\mathcal{F}$ from previous works on question filtering. 

\subsection{Datasets}
\label{subsec:datasets}
We use three academic and one industrial datasets to validate our claims across different data domains and question answering tasks (MR and AS2). 

\mypara{WikiQA:} An AS2 dataset~\cite{yang-etal-2015-wikiqa} with questions from Bing search logs and answer candidates from Wikipedia. We use the most popular setting of training with questions having at least one positive answer candidate, and testing in the \emph{clean} mode with questions having at least one positive and one negative answer candidate. 

\mypara{ASNQ:} A large scale AS2 dataset~\cite{Garg_Vu_Moschitti_2020}~\footnote{\url{https://github.com/alexa/wqa_tanda}} corresponding to Natural Questions (NQ), containing over $60k$ questions and $23M$ answer candidates. Compared to WikiQA, ASNQ has more sophisticated user questions derived from Google search logs and a very high class imbalance ($\sim$ 1 correct in 400 candidate answers) thereby making it a challenging dataset for AS2. We divide the \emph{dev} set from the release of ASNQ into two equal splits with 1336 questions each to be used for validation and testing.

\mypara{SQuAD1.1:} A large scale MR dataset~\cite{rajpurkar-etal-2016-squad}~\footnote{\url{https://rajpurkar.github.io/SQuAD-explorer/}} containing questions asked by crowdworkers with answers derived from Wikipedia articles. Unlike the previous two datasets, SQuAD1.1 requires predicting the exact answer span to answer a question from the provided passage. We divide the \emph{dev} set into two splits of  5266 and 5267 questions for validation and testing respectively. Pr/Re is computed based on exact answer match (EM).

\mypara{AQAD:} A large scale internal industrial dataset containing \textit{non-representative de-identified} user questions from Alexa virtual assistant. Alexa QA Dataset (AQAD) contains 1 million and 50k questions in its train and dev.~sets respectively, with their top answer and confidence scores as provided by the QA system (without any human labels of correctness). Note that the top answer is selected using an answer selection model from hundreds of candidates that are retrieved from a large web-index ($\sim$ 1B web pages). For the purpose of this paper, we use a human annotated portion of AQAD (5k questions other than the train/dev.~splits) as the test split for our experiments. Results on AQAD are presented relative to the baseline {\Mval{0}} due to the data being internal.

\citeauthor{sugawara-etal-2018-makes} previously highlight several shortcomings of using popular MR datasets like SQuAD1.1 for evaluation, due to artifacts such as (i) $35\%$ questions being answerable only using their first 4 tokens, (ii) $76\%$ questions having the correct answer in the sentence with the highest unigram overlap with the question, etc. To ensure that our question filters are learning the capability of the QA system and not these artifacts, we consider datasets from industrial scenarios (where questions are real customer queries) like ASNQ, AQAD \footnote{For ASNQ and IQAD, only $7.04\%$ and $5.82\%$ questions are answered correctly by the highest unigram overlap answer to the question respectively.} and WikiQA in addition to SQuAD.

\subsection{Models}
\label{subsec:models}
For each of the three academic datasets, we use two transformer based models (12 and 24 layer) as $\mathcal{M}$: state-of-the-art RoBERTa-Base and RoBERTa-Large trained with TANDA for WikiQA~\footnotemark[2]~\cite{Garg_Vu_Moschitti_2020}; RoBERTa-Base and RoBERTa-Large-MNLI fine-tuned on ASNQ~\footnotemark[2]~\cite{Garg_Vu_Moschitti_2020}; and, BERT-Base and BERT-Large fine-tuned on SQuAD1.1~\cite{devlin-etal-2019-bert}. For AQAD, we use ELECTRA-Base trained using TANDA~\cite{Garg_Vu_Moschitti_2020} after an initial transfer on ASNQ as $\mathcal{M}$. For the question filter $\mathcal{F}$, we use two different transformer based models (RoBERTa-Base, Large) for each of the four datasets. For WikiQA, ASNQ and SQuAD1.1, the RoBERTa-Base {\F} is used for the 12-layer {\M} and the RoBERTa-Large {\F} is used for the 24-layer {\M}. For AQAD we train both the RoBERTa-Base and RoBERTa-Large {\F} for the single ELECTRA-Base {\M}. All experimental details are presented in Appendix~\ref{app:model_details}, \ref{app:experimental_details} for reproducibility.

\subsection{Baselines}
To demonstrate efficacy of our question filters, we use two question filtering baselines. The first captures well-formedness and intelligibility of questions from a human perspective. For this we train RoBERTa-Base, Large regression models on question well-formedness human annotation scores of the Paralex dataset~\cite{faruqui-das-2018-identifying}~\footnote{\url{https://github.com/google-research-da} \url{tasets/query-wellformedness}}. We denote the resulting filter by {\FW}. For the second baseline, we train a question classifier which predicts whether {\M} will correctly answer a question. This idea has been studied in very recent contemporary works~\cite{varshney2020its,chakravarti2021confident} but for answer verification (not for efficiency). We fine-tune RoBERTa-Base, Large for each dataset to predict whether the target {\M} correctly answers the question or not. We denote this filter by {\FC}.

We exclude comparisons with early exiting strategies~\cite{soldaini-moschitti-2020-cascade,xin-etal-2020-deebert,liu-etal-2020-fastbert} that adaptively reduce the number of transformer layers per sample and aim to improve efficiency of {\M} instead of $\Omega$. Inference batching strategy with multiple samples cannot exploit this efficiency benefit directly, thus these works report efficiency gains through abstract concepts such as FLOPs (Floating Point Operations per Second) using an inference batch-size=1, which is not practical. The efficiency gains from our approach are tangible, since filtering questions can scale down the required number of GPU-compute instances. Furthermore, ideas from these works can easily be combined with ours to add both the efficiency gains to the QA System.

\subsection{Approximating Precision/Recall of {\M}}
\label{subsec:pr_re_curves}

\begin{figure}[t]
\centering
    \begin{subfigure}[t]{\linewidth}
        \begin{subfigure}[t]{0.4825\textwidth}
            \includegraphics[width=1\textwidth]{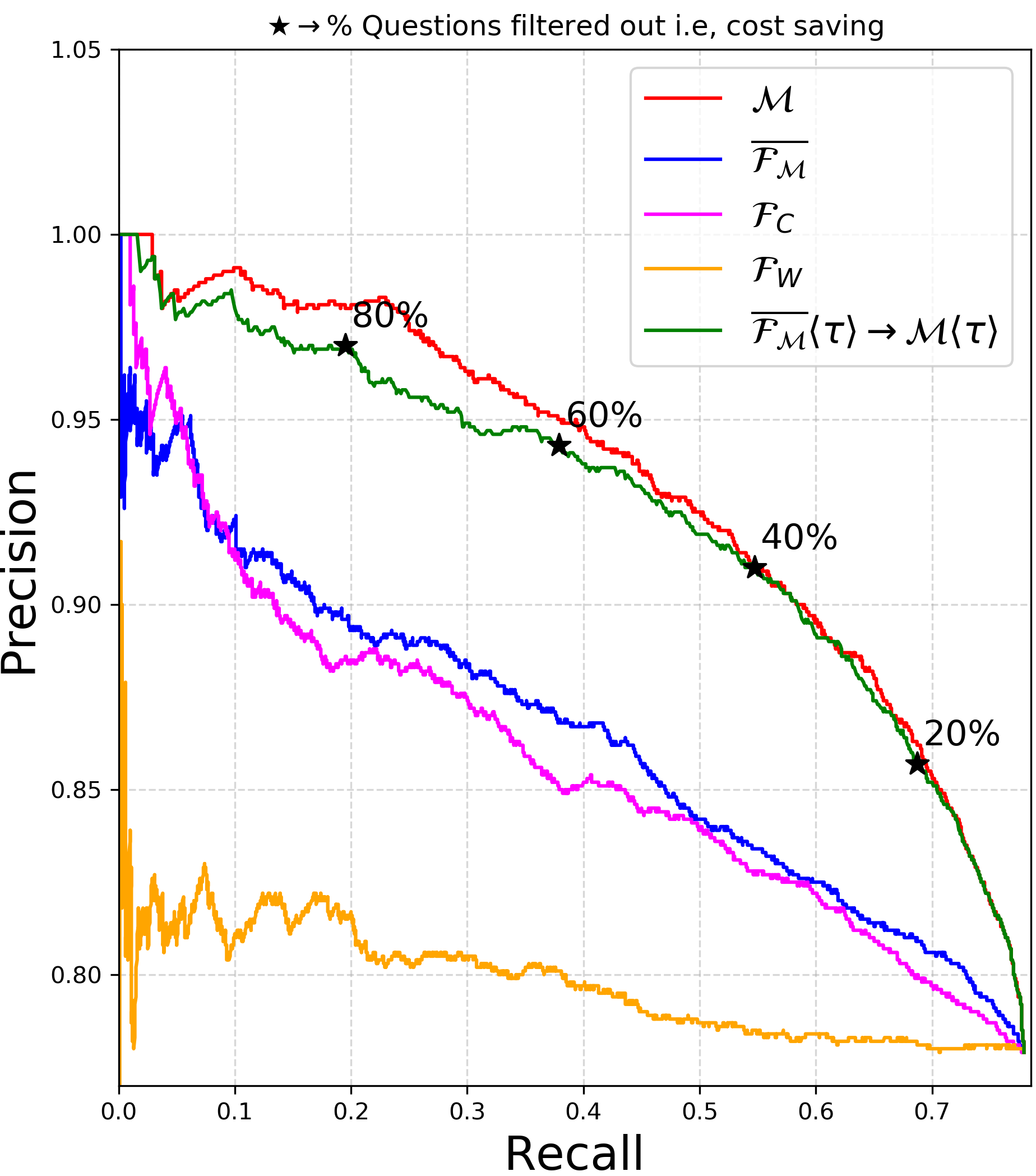}
            \caption{SQuAD 1.1}
        \end{subfigure}
        \begin{subfigure}[t]{0.5025\textwidth}
            \includegraphics[width=1\textwidth]{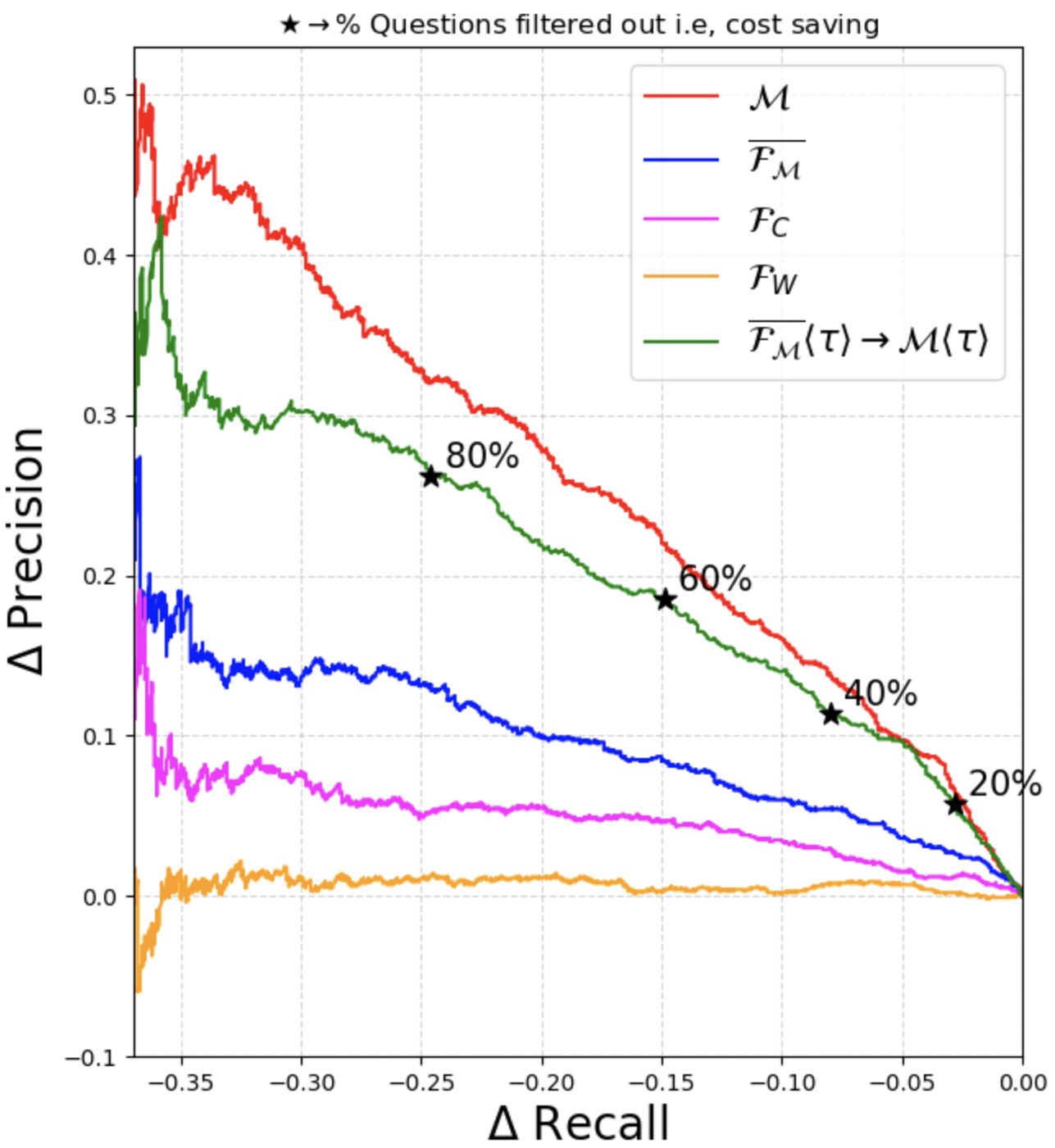}
            \caption{AQAD}
        \end{subfigure}
    \end{subfigure}
    \vspace{-3pt}
    \caption{Pr/Re curves for filters (\textcolor{blue}{\FM}, \textcolor{orange}{\FW}, \textcolor{magenta}{\FC}) and answer model \textcolor{red}{\M} (For AQAD we show $\Delta$Pr/$\Delta$Re w.r.t {\Mval{0}}). Since test splits only contain questions with at least one correct answer, Pr${=}$Re at {\TM}${=}0$.}
    \label{fig:pr_re_graphs}
    \vspace{-3pt}
\end{figure}

Firstly, we want to compare how well our question filter {\F} can approximate the answer score from {\M}. For doing this, we plot the Pr/Re curves of {\M} by varying $\tau_1$ (i.e, ${\mathcal{M}}{\langle\tau_1\rangle}$) and that of filter {\F} by varying $\tau_2$ (i.e, $\mathcal{F}{\langle\tau_2\rangle}$) on the dataset test splits. We consider three options for filter {\F}: our regression head question filter {\FM} and the two baselines: {\FW}, {\FC}. We present graphs on SQuAD1.1 and AQAD using RoBERTa-Base {\F} in Fig.~\ref{fig:pr_re_graphs}. Note that our classification-head filter ($\mathcal{F}_{\mathcal{M}\langle \tau_1 \rangle}$) is trained specific to a particular $\tau_1$ for {\M}, and hence it cannot be directly compared in Fig.~\ref{fig:pr_re_graphs} (since training $\mathcal{F}_{\mathcal{M}\langle \tau_1 \rangle}$ for every $\tau_1 \in [0,1]$ is not feasible). 

The graphs show that {\FM} approximates the Pr/Re of {\M} much better than the baseline filters {\FW} and {\FC}. The gap in approximating Pr/Re of {\M} between {\FM} and {\FC} indicates that learning answer scores is easier than predicting if the model's answer is correct just using the question text.

While these plots independently compare {\FM} and {\M}, in practice, $\widehat{\Omega}$ will operate {\M} at a non-zero threshold $\tau_1$ sequentially after {\FM}${\langle\tau_2\rangle}$ (henceforth we denote this by {\FM}${\langle \tau_2 \rangle}{\rightarrow}${\MTM}). To simplify visualization of the resulting system in Fig.~\ref{fig:pr_re_graphs}, we propose to use a single common threshold {\T} for both {\FM} and {\M}, denoted as {\FM}${\langle \tau \rangle}{\rightarrow}${\MT}. From Fig.~\ref{fig:pr_re_graphs}-(a), (b), the Pr/Re curve for {\FM}${\langle \tau \rangle}{\rightarrow}${\MT} on varying {\T} approximates that of {\M} very well. Using {\FM} however, imparts a large efficiency gain to $\widehat{\Omega}$ as shown by the four operating points that represent the $\%$ of questions filtered out by {\FM}. For example, for AQAD, 60$\%$ of all the questions can be filtered out before running ($\mathcal{R},\mathcal{S},{\mathcal{M}}$) (translating to a cost saving of the same fraction) while only dropping the Recall of $\Omega$ by $3$ to $4$ points. Complete plots having Pr/Re curves for {\FC}${\langle \tau \rangle}{\rightarrow}${\MT} and {\FW}${\langle \tau \rangle}{\rightarrow}${\MT} for all four datasets are included in Appendix~\ref{sec:app_pr_re_graphs}.

\newfloatcommand{capbtabbox}{table}[][1.35\linewidth]

\begin{figure*}
\begin{floatrow}
\capbtabbox{%
    \setlength\extrarowheight{-6pt}
    \resizebox{\linewidth}{!}{
    \begin{tabular}{cccccccccccccc}
    \toprule
    & &   & \multicolumn{5}{c}{\textbf{{\M} : 12-Layer Transformer}}                                                                                                             &   & \multicolumn{5}{c}{\textbf{{\M} : 24-Layer Transformer}}                                                                                                                              \\ 
    & &     & \multicolumn{5}{c}{\textbf{{\F} : RoBERTa-Base }}                                                                                                             &   & \multicolumn{5}{c}{\textbf{{\F} : RoBERTa-Large }}                                                                                                                              \\ \cmidrule{4-8}
    \cmidrule{10-14} 
    \multicolumn{3}{c}{\centering \textbf{Threshold {\TM} for {\M} $\rightarrow$}}    & 0.3 & 0.5 & 0.6 & 0.7 & 0.9  &     & 0.3 & 0.5 & 0.6 & 0.7 & 0.9 \\ \toprule
    \multirow{9}{*}{\begin{turn}{90}\textbf{WikiQA}\end{turn}} & \multirow{2}{*}{{\MTM}} & Pr       	& 84.4  &	87.1  & 88.4  & 88.4  &	97.1  & & 92.1 	& 92.2 	& 92.0	& 92.1  &	95.9          \\ \cmidrule{3-3}
    & & Re       	&  80.2 &  75.3 & 68.7 &  63.0 &  28.0 & &   86.0	&  83.1	&  80.2	&  77.0 &	 57.2          \\ \cmidrule{2-3}
       & \multirow{2}{*}{\small{\FMTM${\langle{\tau_2^*}\rangle}{\rightarrow}${\M}$\langle${\TM}$\rangle$}}         & $\%$ Filter  & 4.1  & 2.9   & 3.7  & 7.0   & 42.8  &  &  0.8  & 3.3  & 6.6  & 7.8  & 18.9  \\ \cmidrule{3-3}
          &   & $\Delta$ Re   & \textcolor{red}{\contour{red}{-}}2.4  &  \textcolor{red}{\contour{red}{-}} 0.8  &  \textcolor{red}{\contour{red}{-}}1.6  &  \textcolor{red}{\contour{red}{-}}2.5  &  \textcolor{red}{\contour{red}{-}} 3.7 &  &   \textcolor{red}{\contour{red}{-}}0.8  &  \textcolor{red}{\contour{red}{-}}2.0  &  \textcolor{red}{\contour{red}{-}} 4.1  &  \textcolor{red}{\contour{red}{-}} 5.0  &  \textcolor{red}{\contour{red}{-}}6.1      \\ \cmidrule{2-3}
        
        & \multirow{2}{*}{\small{\FM${\langle{\tau_2^*}\rangle}{\rightarrow}${\M}$\langle${\TM}$\rangle$}}         & $\%$ Filter  & 4.1    & 17.3     & 20.9    & 21.0     & 44.0   &   &  1.2   & 1.8   & 2.6   & 3.9   & 8.8   \\ \cmidrule{3-3}
        &   & $\Delta$ Re   & \textcolor{red}{\contour{red}{-}}2.8  & \textcolor{red}{\contour{red}{-}}10.3  &  \textcolor{red}{\contour{red}{-}} 9.0  & \textcolor{red}{\contour{red}{-}}8.3  &  \textcolor{red}{\contour{red}{-}}7.4 &   &   \textcolor{red}{\contour{red}{-}}0.8  &  \textcolor{red}{\contour{red}{-}}0.8  &  \textcolor{red}{\contour{red}{-}}0.4  &  \textcolor{red}{\contour{red}{-}}2.1  &  \textcolor{red}{\contour{red}{-}}2.9  \\ \midrule
    \multirow{9}{*}{\begin{turn}{90}\textbf{ASNQ}\end{turn}} & \multirow{2}{*}{\MTM} & Pr                 & 68.2               & 74.6                 &           77.2                & 79.6        & 90.5   &   & 75.7             & 79.6     &         81.9             & 84.5                & 92.9       \\ \cmidrule{3-3}
     &  & Re                &   48.7                &   41.1                &           36.1              &   28.9      &  10.0 &   &   61.1               &   54.4    &          49.3            &   42.7               &  20.7     \\ \cmidrule{2-3}
        & \multirow{2}{*}{\small{\FMTM${\langle{\tau_2^*}\rangle}{\rightarrow}${\M}$\langle${\TM}$\rangle$}}         & $\%$ Filter        & 7.8        & 17.8           &     29.8   & 54.2                 & 83.8    &  & 0.2                              & 10.5    &     15.6         & 29.8     & 66.2      \\ \cmidrule{3-3}
        &  &  $\Delta$ Re         &  \textcolor{red}{\contour{red}{-}}1.8                &  \textcolor{red}{\contour{red}{-}}2.9               &      \textcolor{red}{\contour{red}{-}}4.7  &  \textcolor{red}{\contour{red}{-}}8.2                &   \textcolor{red}{\contour{red}{-}}3.9  &  &   0                              &  \textcolor{red}{\contour{red}{-}} 3.2  &     \textcolor{red}{\contour{red}{-}}3.0           &  \textcolor{red}{\contour{red}{-}}7.4               & \textcolor{red}{\contour{red}{-}}7.0                \\ \cmidrule{2-3}
        
        & \multirow{2}{*}{\small{\FM${\langle{\tau_2^*}\rangle}{\rightarrow}${\M}$\langle${\TM}$\rangle$}}         & $\%$ Filter         &  6.0   & 14.9    & 33.5    & 47.1    & 86.0  &  &  1.0   & 12.1   & 16.1  & 21.6    & 61.6    \\  \cmidrule{3-3}
    &   &  $\Delta$ Re         &   \textcolor{red}{\contour{red}{-}}1.2  &  \textcolor{red}{\contour{red}{-}}2.7  &  \textcolor{red}{\contour{red}{-}}5.3  &  \textcolor{red}{\contour{red}{-}}6.4  &  \textcolor{red}{\contour{red}{-}}5.1 &  &   \textcolor{red}{\contour{red}{-}}0.1  &  \textcolor{red}{\contour{red}{-}}3.3  &   \textcolor{red}{\contour{red}{-}}3.5  &  \textcolor{red}{\contour{red}{-}}3.9  &  \textcolor{red}{\contour{red}{-}}6.3   \\  \midrule
    \multirow{9}{*}{\begin{turn}{90}\textbf{SQuAD 1.1}\end{turn}} & \multirow{2}{*}{{\MTM}} & Pr         &     	82.0    & 	88.7  & 	91.0   & 93.4   & 	96.7   &  & 	86.0 & 	90.1  & 	92.3  & 	94.4  & 	97.6         \\ \cmidrule{3-3}
     &  & Re         &     	  75.0  & 	 63.3 & 	  54.6 &  45.9 & 	 28.7 &  & 	 82.9 & 	 75.3 & 	 67.5 & 	 59.7 & 	 42.4         \\ \cmidrule{2-3}
        & \multirow{2}{*}{\small{\FMTM${\langle{\tau_2^*}\rangle}{\rightarrow}${\M}$\langle${\TM}$\rangle$}}         & $\%$ Filter         & 0   & 9.4    & 12.9   & 27.4  & 61.0    &   &  0  & 2.0    & 6.4  & 14.0  & 	47.5       \\ \cmidrule{3-3}
        &     &  $\Delta$ Re         & 0  & \textcolor{red}{\contour{red}{-}} 2.3  &  \textcolor{red}{\contour{red}{-}}2.4  &  \textcolor{red}{\contour{red}{-}}4.8  &   \textcolor{red}{\contour{red}{-}}7.7  &   &   0  &  \textcolor{red}{\contour{red}{-}}0.5  &  \textcolor{red}{\contour{red}{-}}1.8  &  \textcolor{red}{\contour{red}{-}}3.8 & 	 \textcolor{red}{\contour{red}{-}}3.7               \\ \cmidrule{2-3}    
        & \multirow{2}{*}{\small{\FM${\langle{\tau_2^*}\rangle}{\rightarrow}${\M}$\langle${\TM}$\rangle$}}         & $\%$ Filter            & 1.1   & 9.0    & 14.2    & 36.5    & 63.0   &      & 0   & 2.4  & 5.1  & 18.2   & 41.8   \\ \cmidrule{3-3} 
        &       &  $\Delta$ Re           &  \textcolor{red}{\contour{red}{-}}0.4  &  \textcolor{red}{\contour{red}{-}}2.3  &   \textcolor{red}{\contour{red}{-}}3.2  &  \textcolor{red}{\contour{red}{-}}8.2  &  \textcolor{red}{\contour{red}{-}}8.8 &      &  0 &  \textcolor{red}{\contour{red}{-}}0.5  &  \textcolor{red}{\contour{red}{-}}1.2 &  \textcolor{red}{\contour{red}{-}}5.5  &   \textcolor{red}{\contour{red}{-}}8.1  \\ \midrule 
    \multirow{9}{*}{\begin{turn}{90}\textbf{AQAD} \end{turn}} & \multirow{2}{*}{\MTM} & Pr         &  	{$\uparrow$}9.2  & 	{$\uparrow$}17.9 & 	{$\uparrow$}23.4  & 	{$\uparrow$}28.1  & 	{$\uparrow$}43.0  &  &  	{$\uparrow$}9.2  & 	{$\uparrow$}17.9  & 	{$\uparrow$}23.4  & 	{$\uparrow$}28.1 & 	{$\uparrow$}43.0     \\ \cmidrule{3-3}
    &  &  Re         &  	 {$\downarrow$}4.5 & 	 {$\downarrow$}12.1 & 	 {$\downarrow$}15.7 & 	 {$\downarrow$}20.1 & 	 {$\downarrow$}31.9 &  &  	 {$\downarrow$}4.5 & 	 {$\downarrow$}12.1 & 	 {$\downarrow$}15.7 & 	 {$\downarrow$}20.1 & 	 {$\downarrow$}31.9     \\ \cmidrule{2-3}
        & \multirow{2}{*}{\small{\FMTM${\langle{\tau_2^*}\rangle}{\rightarrow}${\M}$\langle${\TM}$\rangle$}}         & $\%$ Filter            & 20.8    & 43.2    & 53.9    & 65.2    & 89.4   & &  21.9    & 45.8    & 56.9    & 66.2   & 89.9   \\ \cmidrule{3-3}
        &  & $\Delta$ Re           &  \textcolor{red}{\contour{red}{-}}3.4  &   \textcolor{red}{\contour{red}{-}}5.0  &   \textcolor{red}{\contour{red}{-}}5.6  &   \textcolor{red}{\contour{red}{-}}6.0  &   \textcolor{red}{\contour{red}{-}}3.4 & &    \textcolor{red}{\contour{red}{-}}3.2  &   \textcolor{red}{\contour{red}{-}}4.9  &  \textcolor{red}{\contour{red}{-}}5.5  &   \textcolor{red}{\contour{red}{-}}4.6  &  \textcolor{red}{\contour{red}{-}}3.0  \\ \cmidrule{2-3}    
        &\multirow{2}{*}{\small{\FM${\langle{\tau_2^*}\rangle}{\rightarrow}${\M}$\langle${\TM}$\rangle$}}  & $\%$ Filter   & 17.8   & 48.2    & 59.8    & 75.7    & 91.7   & & 15.2    & 48.7    & 57.3    & 72.0    & 93.8    \\ \cmidrule{3-3} 
        &  &  $\Delta$ Re          &  \textcolor{red}{\contour{red}{-}}2.8  &   \textcolor{red}{\contour{red}{-}}6.6  &   \textcolor{red}{\contour{red}{-}}7.5  &  \textcolor{red}{\contour{red}{-}}8.7  &   \textcolor{red}{\contour{red}{-}}3.8 & &   \textcolor{red}{\contour{red}{-}}1.7  &  \textcolor{red}{\contour{red}{-}}5.7  &  \textcolor{red}{\contour{red}{-}}6.2  &  \textcolor{red}{\contour{red}{-}}7.0  &   \textcolor{red}{\contour{red}{-}}3.9  \\ \bottomrule
    \end{tabular}
    }
}{%
    \caption{Filtering gains and drop in recall for question filters operating at optimal filtering threshold {\TFBest}. For a particular filter {\F} operating with answer model {\MTM}, $\Delta$ Re refers to the difference in Recall of {\F}${\langle{\tau_2}^*\rangle}{\rightarrow}${\MTM} and {\MTM}. $\%$ Filter refers to the $\%$ of questions preemptively discarded by {\F}. {\MTM} results for AQAD are relative to {\M}$\langle 0 \rangle$.}
    \label{tab:main_table_small} 
}
\centering
    \ffigbox[0.6\linewidth]
    {
    \begin{subfigure}[t]{0.95\linewidth}
        \includegraphics[width=0.87\linewidth]{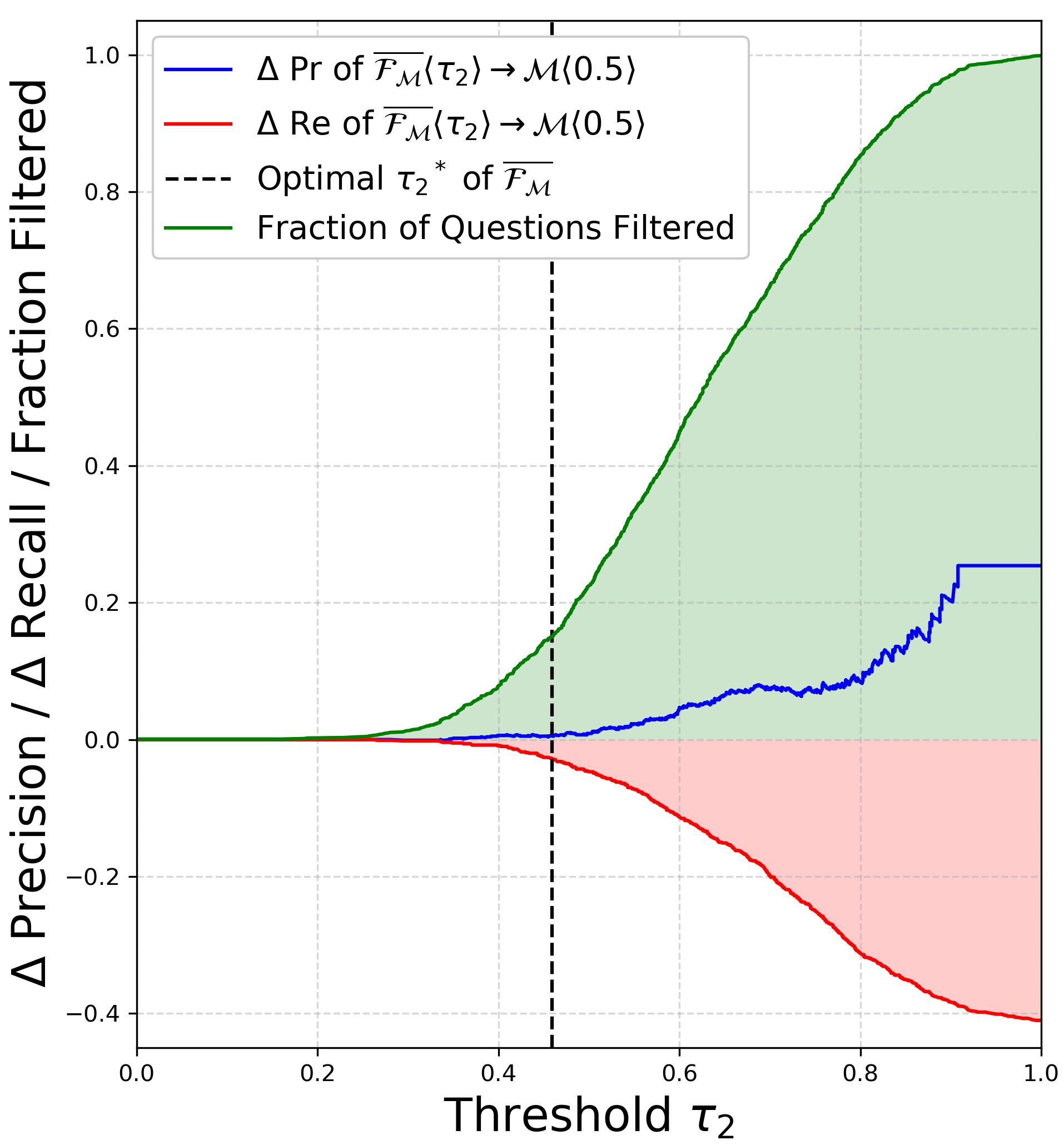}
        \vspace{-0.15cm}
        \caption{{\small ASNQ}}
    \end{subfigure}
    \begin{subfigure}[t]{0.95\linewidth}
        \includegraphics[width=0.87\linewidth]{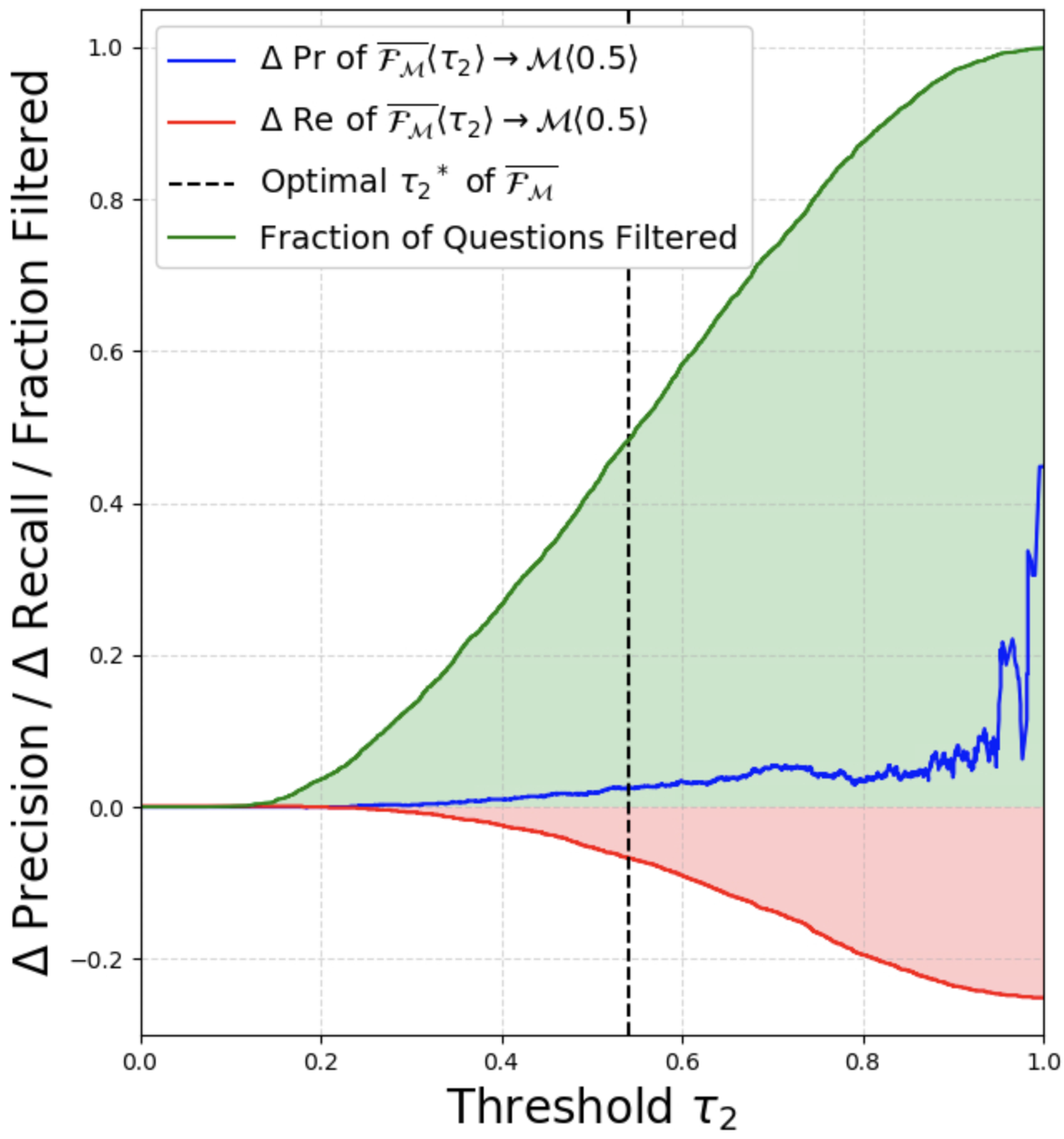}
        \vspace{-0.15cm}
        \caption{{\small AQAD}}
    \end{subfigure}
    }
    {
    \vspace{-0.25cm}
  \caption{$\Delta$ Pr/$\Delta$ Re plots and fraction of questions filtered on varying {\TF} for RoBERTa-Base {\FM}${\langle\tau_2\rangle}{\rightarrow}${\M}$\langle 0.5 \rangle$ on ASNQ and AQAD datasets.}
    \label{fig:threshold_vary_plots}
}
\end{floatrow}
\end{figure*}

\subsection{Selecting Threshold {\TF} for {\F}}
\label{sec:select_threshold_for_F}
When adding a question filter {\F} to $\Omega(\mathcal{R},\mathcal{S},{\mathcal{M}})$, the operating threshold {\TF} of {\F} is a user-tunable parameter which can be varied per the use-case: efficiency desired at the expense of recall. This user-tunable {\TF} is a prominent advantage of our approach since one can decide what fraction of questions to filter out based on how much recall one can afford to drop. We plot the variation of the fraction of questions filtered by {\F} along with the change in Pr/Re of $\widehat{\Omega}$ on varying {\TF} in Fig.~\ref{fig:threshold_vary_plots}. Specifically we consider the ASNQ and AQAD datasets, and {\M} operating at {\TM}${=}0.5$. From Fig.~\ref{fig:threshold_vary_plots}(a) we can observe that for ASNQ, our filter {\FM} can obtain ${\sim}18\%$ filtering gains while only losing a recall of ${\sim}3$ points. {\FM} can obtain even better filtering gains on AQAD: from Fig.~\ref{fig:threshold_vary_plots}(b) ${\sim}40\%$ filtering by only losing ${\sim}4$ points of recall. Complete plots for all datasets can be found in Appendix~\ref{sec:app_threshold_vary_plots}.

We now present one possible way to choose an operating threshold {\TF} for filter {\F}. For a QA system $\widehat{\Omega}(\mathcal{F}\langle\tau_2\rangle,\mathcal{R},\mathcal{S},${\MTM}$)$, we find the threshold $\tau_2^*$ for {\F} at which it best approximates the answering/abstaining choice of {\MTM}. Specifically, we use the dev.~split of the datasets to find {\TFBest}$\in [0,1]$ such that $\mathcal{F}\langle\tau_2^*\rangle$ obtains the highest F1-score corresponding to the binary decision of answering or abstaining by {\MTM}. We present empirical results of our filters at different thresholds {\TM} of {\M} in Table~\ref{tab:main_table_small}. We evaluate the $\%$ of questions filtered out by {\F}$\langle{\tau_2^*}\rangle$ (efficiency gains) and the resulting drop in recall of {\F}${\langle{\tau_2^*}\rangle}{\rightarrow}${\M}$\langle${\TM}$\rangle$ from {\MTM} on the test split of the dataset. For each dataset and model {\M}, we train one regression head filter {\FM} and five classification head filters {\FMTM}: one at every threshold {\TM} for {\M} $\in \{0.3,0.5,0.6,0.7,0.9\}$. For regression head {\FM}, the optimal {\TFBest} is calculated independently for every {\TM} of {\M}.

\mypara{Results:} From Table~\ref{tab:main_table_small} we observe that our question filters (both with classification and regression heads) can impart filtering efficiency gains (of different proportions) while only incurring a small drop in recall.
For example on ASNQ (12-layer {\M, \F}), {\FMTval{0.5}} is able to filter out $17.8\%$ of the questions while only incurring a drop in recall of $2.9\%$. On ASNQ (24-layer {\M, \F}), {\FM} is able to filter out $21.6\%$  of the questions with a drop in recall of only $3.9\%$ at {\TM}${=}0.7$. Barring some cases at higher thresholds, {\FM} achieves comparable filtering performance to {\FMTM}. The best filtering gains are obtained on the industrial AQAD dataset having real world noise, where for {\TM}$=0.5$, the 24-layer {\FMTval{0.5}} can filter out $45.8\%$ of all questions only incurring a drop in recall of $4.9\%$.

We observe that the filtering gains at the optimal {\TFBest} are inversely correlated with the precision of {\M}. For example, for (12-layer {\M, \F}) at {\TM}$=0.5$, the Pr of SQuAD 1.1 and ASNQ is 88.7 and 74.6 respectively, and that of AQAD is significantly lower than ASNQ (due to real world noise). The $\%$ of questions filtered by {\FMTM}${\langle\tau_2^*\rangle}$ or {\FM}${\langle\tau_2^*\rangle}$ increases in the order from $9{-}9.4\%$ to $14.9{-}17.8\%$ to $43.2{-}48.2\%$. The efficiency gain of our filters thus increases as the QA task becomes increasingly difficult (SQuAD 1.1 $\rightarrow$ ASNQ $\rightarrow$ AQAD). Furthermore, except for some cases on WikiQA, we observe that our question filters increase the precision of the system (for full table with $\Delta$Pr and $\Delta$F1 refer Table~\ref{tab:main_table} in Appendix~\ref{sec:app_optimal_F_threshold}). This is in line with our observations in Fig.~\ref{fig:calibration_plots} and \citeauthor{kamath-etal-2020-selective}.

WikiQA (873 questions) is a very small dataset for efficiently distilling information from a transformer {\M}. Standard distillation~\cite{HinVin15Distilling} often requires millions of training samples for efficient learning. To mitigate this, we extrapolate sequential fine-tuning as presented in \cite{Garg_Vu_Moschitti_2020} for learning question filters for WikiQA. 
We perform a two step learning of {\FM}, {\FMTM}: first on ASNQ and then on WikiQA. The results for WikiQA in Table~\ref{tab:main_table_small} correspond to this paradigm of training {\FM} and {\FMTM}, and demonstrate that our approach works to a reasonable level even on very small datasets. This also has implication towards shared semantics of question filters for models {\M} trained on different datasets.

The drop in Re and filtering gains are contingent on the Pr/Re curve of {\M} for the dataset. At higher thresholds (say {\TM}${=}0.9$), if the drop in recall due to {\FMTM} or {\FM} at {\TFBest} is more than desirable, then one can reduce the value of {\TF} down from {\TFBest} by reducing the efficiency gains using plots like Fig.~\ref{fig:threshold_vary_plots}.

\mypara{Comparison with Baselines:} We also present results on optimal {\TFBest} for {\FW} and {\FC} in Table~\ref{tab:baselines} for ASNQ (complete results for all datasets are in Appendix~\ref{sec:app_optimal_F_threshold}). When compared with the performance of {\FM} and {\FMTM} for ASNQ in Table~\ref{tab:main_table_small}, both {\FW} and {\FC} perform inferior in terms of filtering performance. {\FW}, which evaluates well-formedness of the questions from a human perspective, is unable to filter out any questions even when operating at its optimal threshold. This indicates that human-supervised filtering of ill-formed questions is sub-optimal from an efficiency perspective. {\FC} gets better performance than {\FW}, but always trails {\FM} and {\FMT} either in terms of a smaller $\%$ of questions filtered or a larger drop in recall incurred.

\mypara{Efficiency Gains from Filtering:} Under simplifying assumptions, the computational resources required to answer questions within a fixed time budget over a fixed set of documents scales roughly linearly with the number of concurrent requests that need to be processed by $\Omega$.
We present a simple analysis on ASNQ (ignoring cost of retrieval $\mathcal{R},\mathcal{S}$) considering 1000 questions (ASNQ has~400 candidate answers/question) and a batch-size${=}100$. {\M} requires $1000*400/100{=}4000$ transformer forward passes ($\text{max}\_\text{seq}\_\text{length}{=}128$, standard for QA tasks due to long answers). On the other hand, $\text{max}\_\text{seq}\_\text{length}{=}32$ suffices for {\F}. Since inference latency of transformers scales roughly quadratic over input sequence length, 1 batch through {\M} is $4^2{=}16$ times slower than through {\F}. Assuming 20\% question filtering by {\F}, {\M} now only answers 800 questions (3200 forward passes of M), while adding $1000/100{=}10$ forward passes of {\F}. The \%-cost reduction in time is $19.968\%{\sim}20\%$. We perform inference on ASNQ test-set on one V100-GPU (with {\FM} set to filter out 20\% as per above) and observe latency dropping from $531.29$s $\rightarrow$ $433.16$s ($18.47\%$, slightly lower than calculated $19.968\%$ due to input/output overheads). The latency reduction can also translate to a reduction in the number of GPU compute resources required when performing inference in parallel. Furthermore, in practice, our filter will also provide cost/latency savings by not performing document retrieval for the filtered out questions.

\begin{table}
\centering
\resizebox{\linewidth}{!}{
\begin{tabular}{ccccccc}
\toprule
\multirow{2}{*}{\textbf{{\TM} of {\M} $\downarrow$}}  & & \multicolumn{2}{c}{{\FW}} & &\multicolumn{2}{c}{{\FC}} \\ \cmidrule{3-4} \cmidrule{6-7}
& & Base & Large & & Base & Large \\
\midrule
0.3 & & 0.3 / \textcolor{red}{\contour{red}{-}}0.1 & 0.1 / \textcolor{red}{\contour{red}{-}}0.2 & & 	0.2 / \textcolor{red}{\contour{red}{-}}0.1 & 0.3 / \textcolor{red}{\contour{red}{-}}0.2 \\

0.5 & & 0.9 / \textcolor{red}{\contour{red}{-}}0.4 & 0.1 / \textcolor{red}{\contour{red}{-}}0.1 && 1.0 / \textcolor{red}{\contour{red}{-}}0.2 & 0.5 / \textcolor{red}{\contour{red}{-}}0.3 \\

0.6 && 0.9 / \textcolor{red}{\contour{red}{-}}0.4 & 0.1 / \textcolor{red}{\contour{red}{-}}0.2 && 22.3 / \textcolor{red}{\contour{red}{-}}5.0 & 2.7 / \textcolor{red}{\contour{red}{-}}0.8 \\

0.7 && 1.0 / \textcolor{red}{\contour{red}{-}}0.2 & 0.2 / \textcolor{red}{\contour{red}{-}}0.1 && 38.5 / \textcolor{red}{\contour{red}{-}}6.2 & 7.2 / \textcolor{red}{\contour{red}{-}}1.3 \\

0.9 && 0 / 0 & 0 / 0 && 84.4 / \textcolor{red}{\contour{red}{-}}4.8 & 62.5 / \textcolor{red}{\contour{red}{-}}8.1 \\ \bottomrule
\end{tabular}
}
\vspace{-8pt}
\caption{$\%$ Filter / $\Delta$ Recall results of baseline question filters {\FW} and {\FC} at optimal {\TFBest} on ASNQ. Metrics are similar to those in Table~\ref{tab:main_table_small}. Base and Large refers to RoBERTa-Base and RoBERTa-Large question filters.} 
\label{tab:baselines}
\end{table}

\subsection{Qualitative Analysis}
In Table~\ref{tab:error_analysis}, we discuss some examples to highlight a few shortcomings of {\F}. Both {\F,\M} can successfully filter out non-factual queries asking for opinions (examples 1, 2).  Identifying popular entities like ("Jennifer Lopez", "Lakers", "horses") while training, {\F} incorrectly assumes that a question composed of these entities will be answered by the system. While it may happen that due to unavailability of a web document having the exact answer, {\M} might not answer the question (examples 3, 4). On the other hand, being unfamiliar with entities not encountered during training ("Ahsoka Tano","Mandalorian") or syntactically-complex questions, {\F} preemptively might filter out questions which actually will be answered by {\M} (examples 5, 6).

\section{Conclusion and Future Work}
\label{con}
\vspace{-5pt}
In this paper, we have presented a novel paradigm of training a question filter to capture the semantics of a QA system's answering capability by distilling the knowledge of the answer scores from it. Our experiments on three academic and one industrial QA benchmark show that the trained question models can estimate the Pr/Re curves of the QA system well, and can be used to effectively filter questions while only incurring a small drop in recall. 

\begin{table}[t]
\centering
\resizebox{\linewidth}{!}{
\begin{tabular}{ccc}
\toprule
\textbf{Question} & {\F} & {\M} \\ \midrule
1. What's your favorite movie series? & \textcolor{red}{\xmark} & \textcolor{red}{\xmark} \\
2. Where is the key to the building? & \textcolor{red}{\xmark} & \textcolor{red}{\xmark} \\
3. Was Jennifer Lopez a cheerleader for the Lakers? & \textcolor{darkgreen}{\cmark} & \textcolor{red}{\xmark} \\
4. What are two things that all horses have? & \textcolor{darkgreen}{\cmark} & \textcolor{red}{\xmark} \\
5. Which mayors of New York City had the name David? & \textcolor{red}{\xmark} & \textcolor{darkgreen}{\cmark} \\
6. Does Ahsoka Tano appear in the Mandalorian? & \textcolor{red}{\xmark} & \textcolor{darkgreen}{\cmark} \\
\bottomrule
\end{tabular}
}
\vspace{-8pt}
\caption{Qualitative examples of questions along with the filtering decision of {\M}$\langle 0.5 \rangle$ and {\FM}$\langle 0.5 \rangle$ (\textcolor{darkgreen}{\cmark} / \textcolor{red}{\xmark} indicates clearing / failing the threshold).} 
\label{tab:error_analysis}
\end{table}

An interesting future work direction is to analyze the impact/behavior of the question filters in a cross-domain setting, where the training and testing corpora are from different domains. This would allow examining the transferability of the semantics learned by the question filters. A complementary future work direction could be knowledge distillation from a sophisticated answer verification module like~\cite{DBLP:journals/corr/abs-1904-04792,kamath-etal-2020-selective,zhang-etal-2021-knowing}.

In addition to providing efficiency gains, the question filters could be used to qualitatively study the characteristics of questions that are likely to lead to low answer confidence scores. This can (i) help error analysis for improving the accuracy of QA systems, and (ii) be used for efficient sampling of training questions that are harder to be answered by the target QA system.

Our approach for training the question filters proposes the idea of partial-input knowledge distillation (e.g., using only questions instead of QA pairs). This concept can possibly be extended to other NLP problems for achieving compute efficiency gains, improved explainability (e.g., to what extent a partial-input influences model prediction) and qualitative analysis.

\section*{Acknowledgements}
We thank the anonymous reviewers and meta-reviewer for their valuable suggestions. We thank Thuy Vu for developing and sharing the human annotated data used in the AQAD dataset.

\bibliographystyle{acl_natbib}
\bibliography{references}

\clearpage

\appendix
\section*{Appendix}

\section{Dataset Details}
\label{app:datasets}
All the datasets considered in this paper are in the English language. The dataset statistics for all the datasets are presented in Table~\ref{tab:datasets}.

\mypara{WikiQA:} A small-scale answer sentence selection dataset released by ~\citeauthor{yang-etal-2015-wikiqa} where the candidate answers are extracted from Wikipedia and the questions are derived from query logs of the Bing search engine. The search engine used for retrieving candidate answers is the Microsoft Bing web search engine. This dataset can be downloaded from the provided link~\footnote{\url{http://aka.ms/WikiQA}}. This dataset has a subset of questions having no correct answer sentence (\emph{all$-$}) or have only correct answer sentences (\emph{all$+$}). The training is done by removing \emph{all$-$} questions, and the testing is done by removing both the \emph{all$-$} and \emph{all$+$} questions. 

\mypara{ASNQ:} A large-scale answer sentence selection dataset released by ~\citeauthor{Garg_Vu_Moschitti_2020} where the candidate answers are from Wikipedia pages and the questions are from search queries of the Google search engine. ASNQ is a modified version of the Natural Questions (NQ)~\cite{Kwiatkowski_NQ} dataset by converting it from a MR dataset to an AS2 dataset. This is done by labelling sentences from the long answers which contain the short answer string as positive \emph{correct} answer candidates and all others as negatives. This dataset can be downloaded from the provided link~\footnote{\url{https://github.com/alexa/wqa_tanda}}. The dev split provided in the link is randomly divided into two equal components having 1336 questions each: one for validation, and the other for testing.

\mypara{SQuAD1.1:} A large-scale machine reading dataset released by ~\citeauthor{rajpurkar-etal-2016-squad} where the questions have been written by crowdworkers and the answers are derived from Wikipedia articles. This requires predicting the start and end position (exact span) of the answer for a question from within the associated passage. Due to the hidden test set of SQuAD1.1 which is used for the leaderboard, we randomly divided the dev split into two components: one having 5266 questions (to be used for validation), and the other having 5267 questions (to be used for testing). All results on SQuAD in the paper are reported considering exact answer match. This dataset can be found at this link~\footnote{\url{https://rajpurkar.github.io/SQuAD-explorer/}}. 

\mypara{AQAD:} A large scale internal industrial QA dataset derived from Alexa virtual assistant. Alexa QA Dataset (AQAD) contains 1 million and 50k questions in its train and dev.~sets respectively, with their top answer and confidence scores as provided by the QA system. Note that the question answer pairs are without any human labels of correctness/incorrectness. The top answer is selected using an answer sentence selection model from hundreds of candidates that are retrieved from a large web-index ($\sim$ 1B web pages). For testing, we use 5000 questions (other than those in the train/dev.~splits), each of which is human annotated with a label corresponding to the top answer from the QA system being correctly or incorrect. For learning the correctness filter {\FC} baseline on AQAD, we use an additional annotated split of 2,500 questions other than the train/dev./test splits.

\begin{table}
\centering
\resizebox{0.9\linewidth}{!}{
\begin{tabular}{cccc}
\toprule
\textbf{Dataset} & Train & Validation & Test \\ \midrule
WikiQA & 873 & 121 & 237 \\
ASNQ & 57,242 & 1,336 & 1,336 \\
SQuAD 1.1 & 87,342 & 5,266 & 5,267 \\
AQAD & 1,000,000 & 50,000 & 5,000 \\
\bottomrule
\end{tabular}
}
\vspace{-5pt}
\caption{Dataset statistics providing exact details of number of questions in the train/validation/test split.} 
\label{tab:datasets}
\end{table}

\section{Model Details}
\label{app:model_details}
For each of the datasets we describe the details of the answer models {\M} for reproducibility purposes:

\begin{figure*}[t]
\centering

    \begin{subfigure}[t]{\textwidth}
        \begin{subfigure}[t]{0.24\textwidth}
            \includegraphics[width=1\textwidth]{Plots/Experiments/Pr_Re_Curves/squad.png}
            \caption{SQuAD 1.1}
        \end{subfigure}
        \begin{subfigure}[t]{0.25\textwidth}
            \includegraphics[width=1\textwidth]{Plots/Experiments/Pr_Re_Curves/iqad.png}
            \caption{AQAD}
        \end{subfigure}
        \begin{subfigure}[t]{0.24\textwidth}
            \includegraphics[width=1\textwidth]{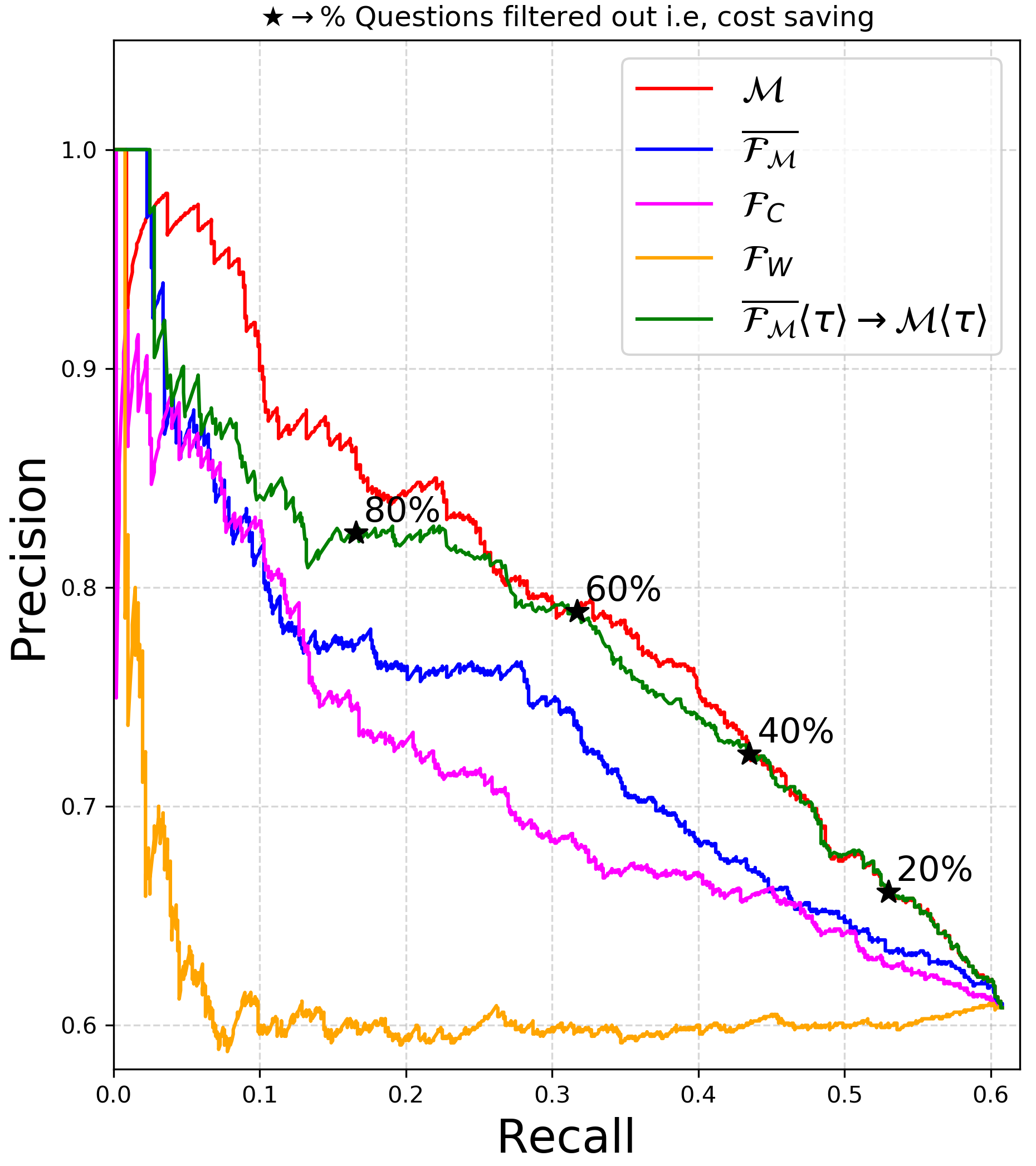}
            \caption{ASNQ}
        \end{subfigure}
        \begin{subfigure}[t]{0.24\textwidth}
            \includegraphics[width=1\textwidth]{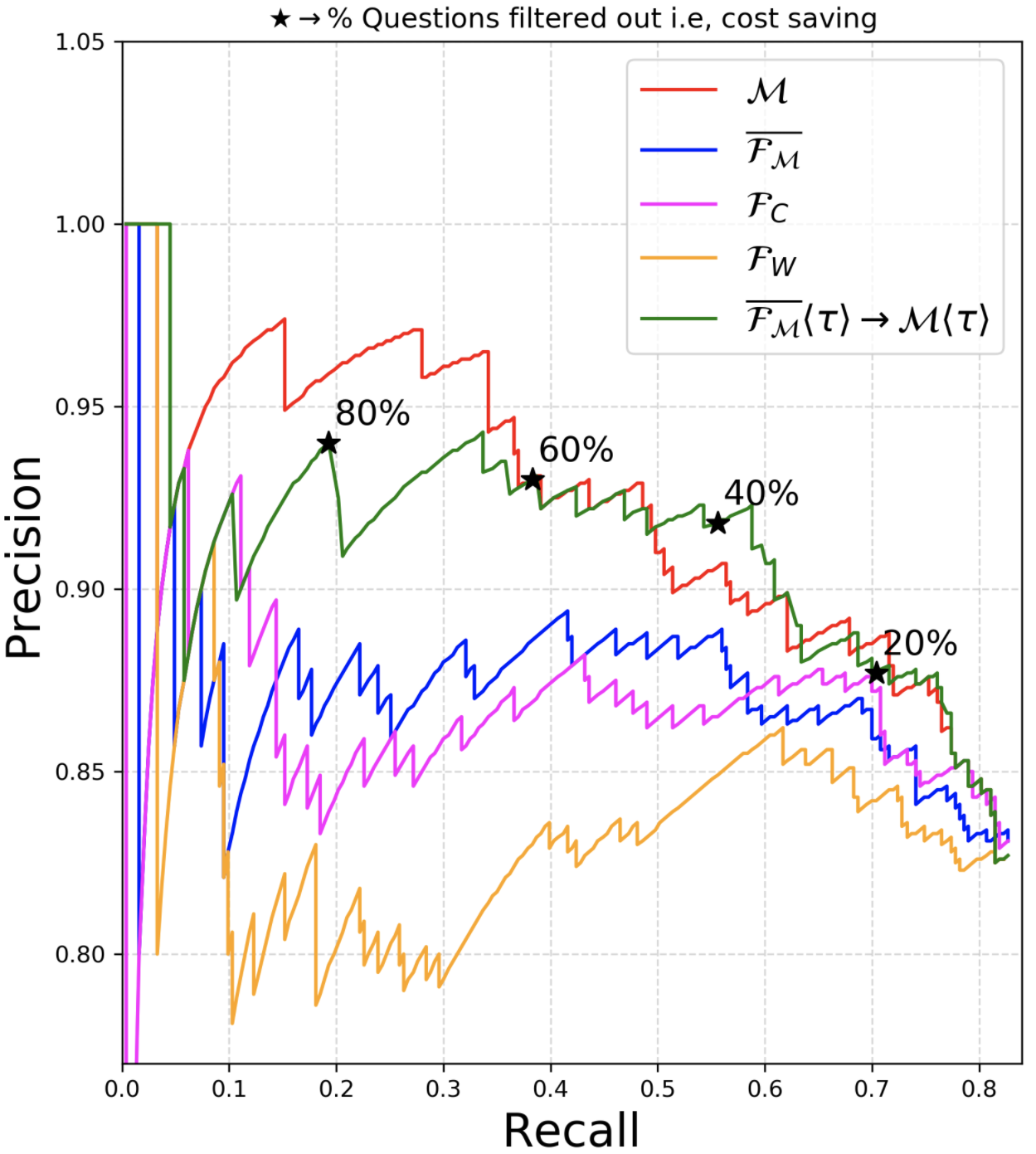}
            \caption{WikiQA}
        \end{subfigure}
    \end{subfigure}
    
    \begin{subfigure}[t]{\textwidth}
        \begin{subfigure}[t]{0.24\textwidth}
            \includegraphics[width=1\textwidth]{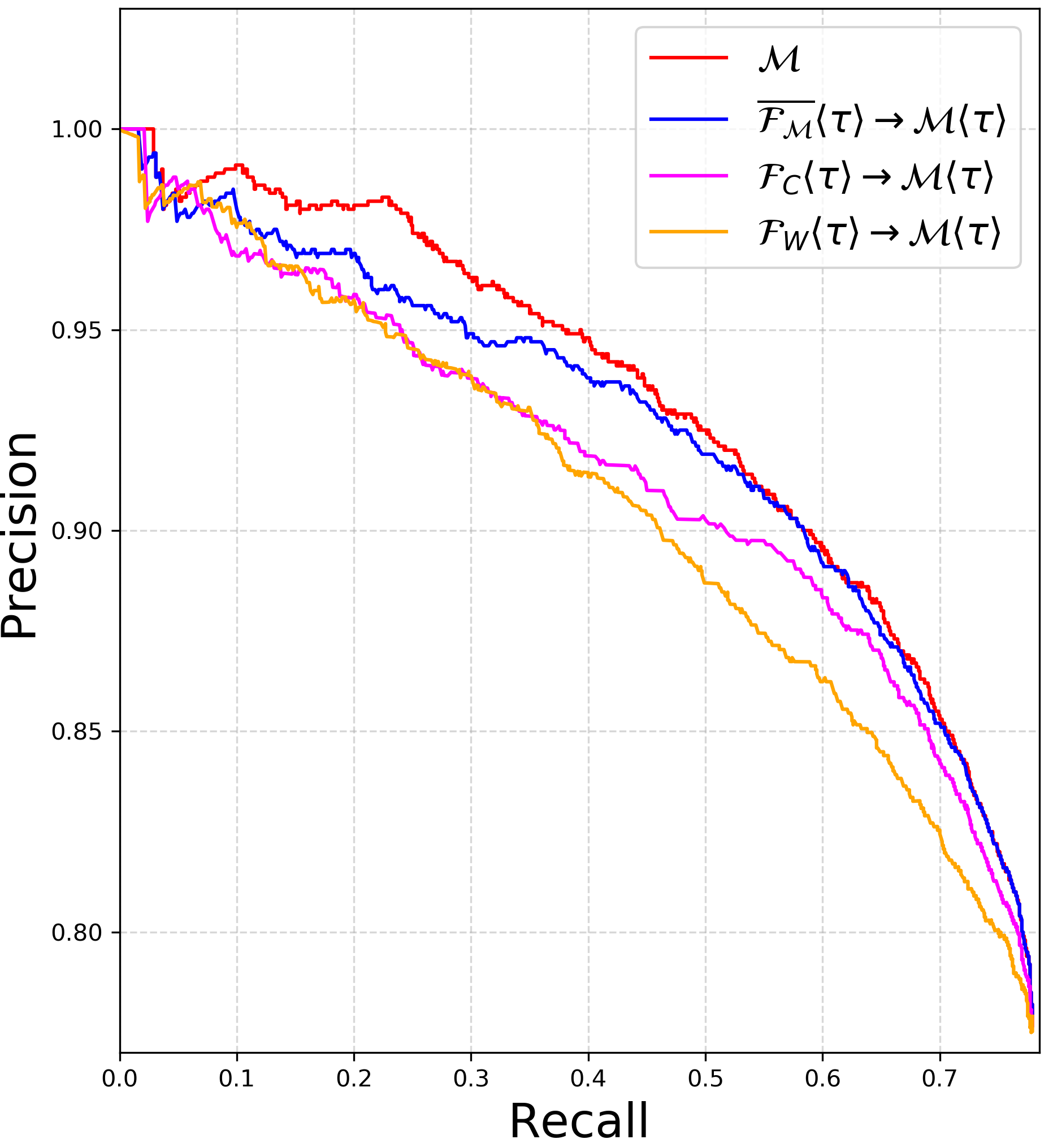}
            \caption{SQuAD 1.1}
        \end{subfigure}
        \begin{subfigure}[t]{0.25\textwidth}
            \includegraphics[width=1\textwidth]{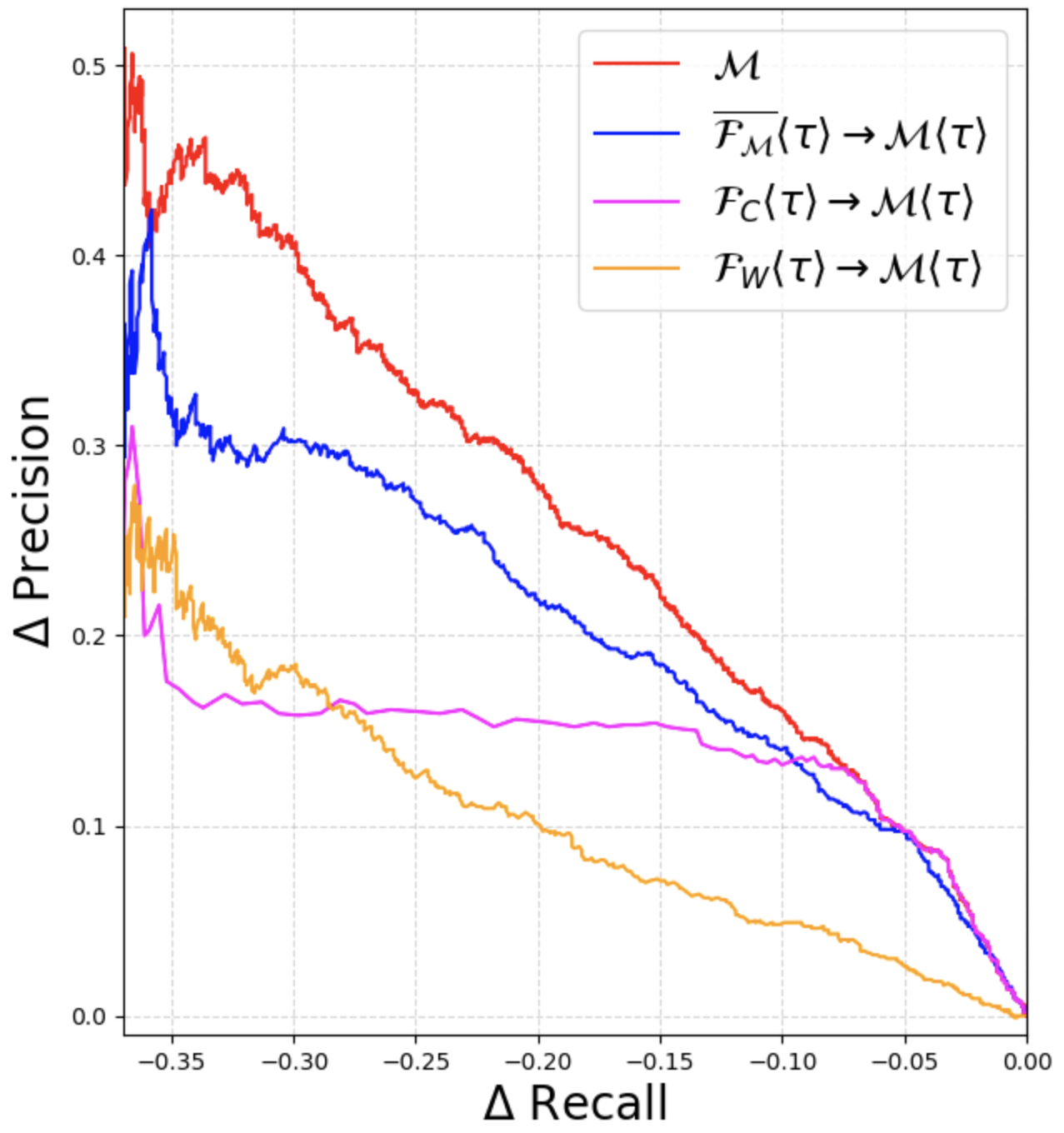}
            \caption{AQAD}
        \end{subfigure}
        \begin{subfigure}[t]{0.24\textwidth}
            \includegraphics[width=1\textwidth]{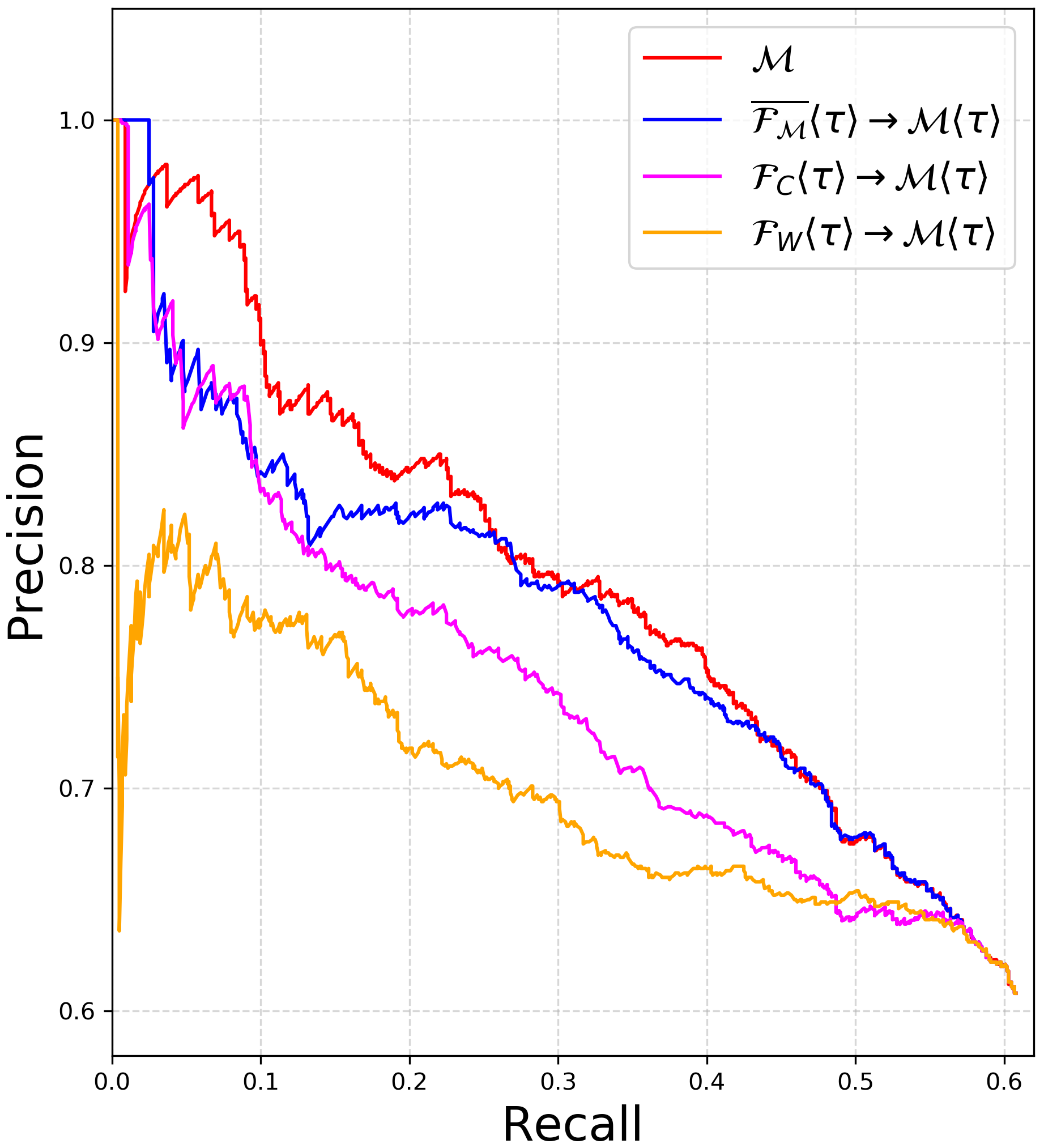}
            \caption{ASNQ}
        \end{subfigure}
        \begin{subfigure}[t]{0.24\textwidth}
            \includegraphics[width=1\textwidth]{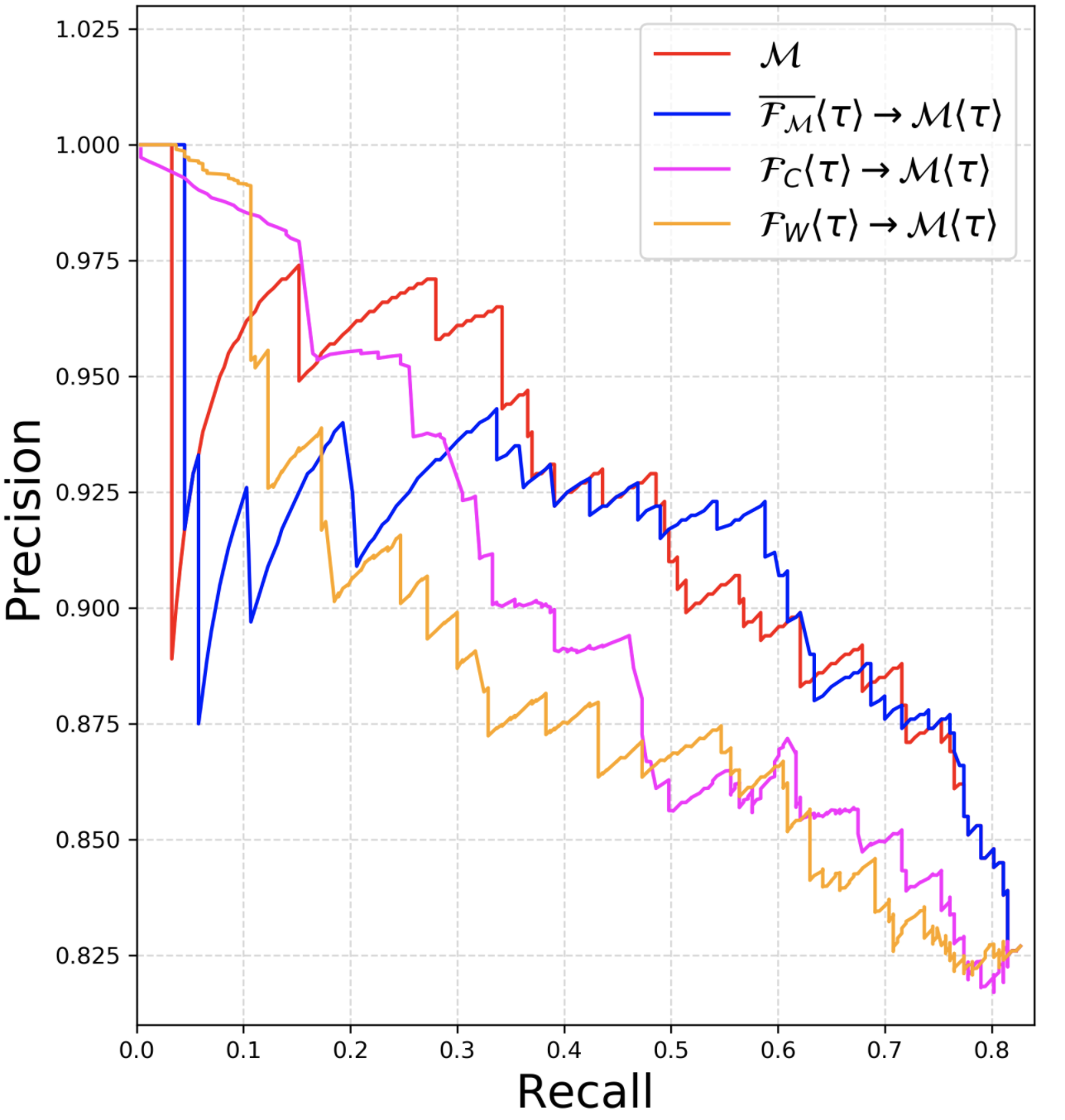}
            \caption{WikiQA}
        \end{subfigure}
    \end{subfigure}
    
    \caption{Pr/Re curves for filters (\textcolor{blue}{\FM}, \textcolor{orange}{\FW}, \textcolor{magenta}{\FC}) and answer model \textcolor{red}{\M} are presented in (a)-(d). Pr/Re curves for filters jointly operating with {\M}, i.e, \textcolor{blue}{\FM${\langle \tau \rangle}{\rightarrow}${\MT}}, \textcolor{orange}{\FW${\langle \tau \rangle}{\rightarrow}${\MT}}, \textcolor{magenta}{\FC${\langle \tau \rangle}{\rightarrow}${\MT}} are presented in (e)-(h). For AQAD we show $\Delta$Pr/$\Delta$Re w.r.t {\Mval{0}}.}
    \label{fig:app_pr_re_graphs}
\end{figure*}

\begin{figure*}[t]
\centering    
    
    \begin{subfigure}[t]{\textwidth}
        \begin{subfigure}[t]{0.24\textwidth}
            \includegraphics[width=1\textwidth]{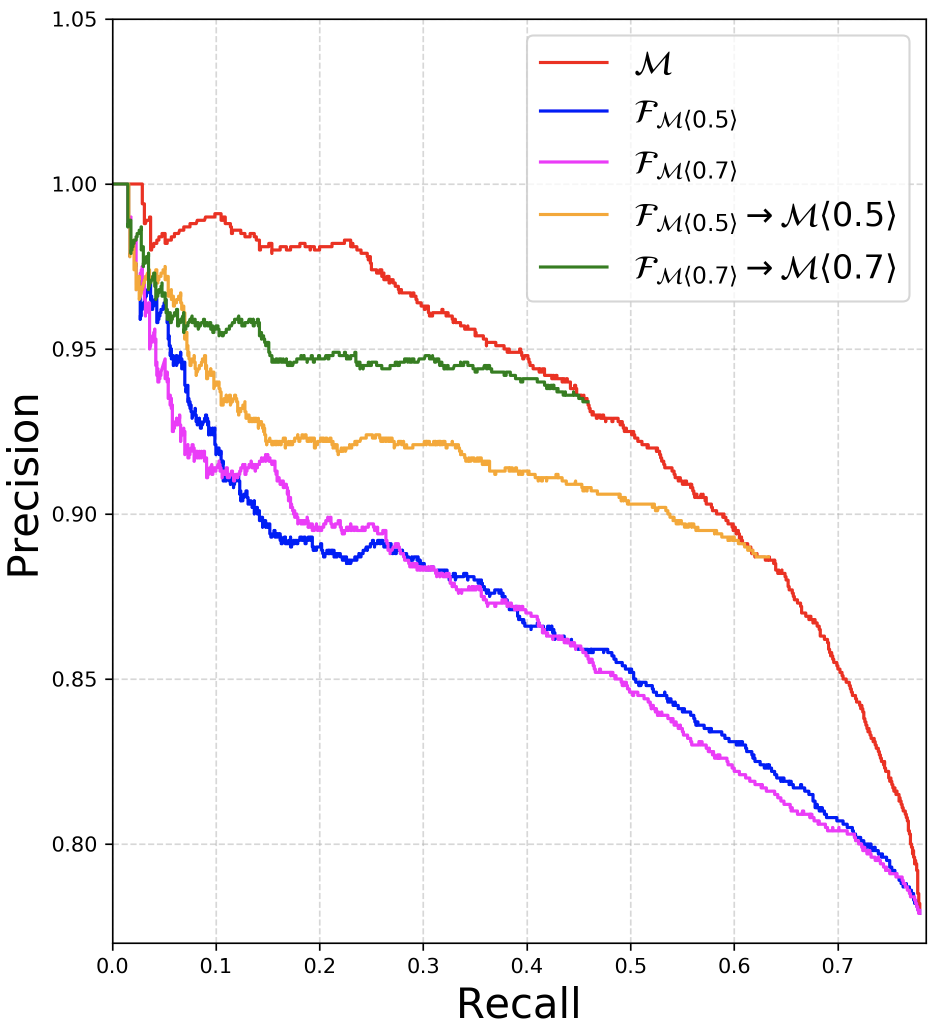}
            \caption{SQuAD 1.1}
        \end{subfigure}
        \begin{subfigure}[t]{0.25\textwidth}
            \includegraphics[width=1\textwidth]{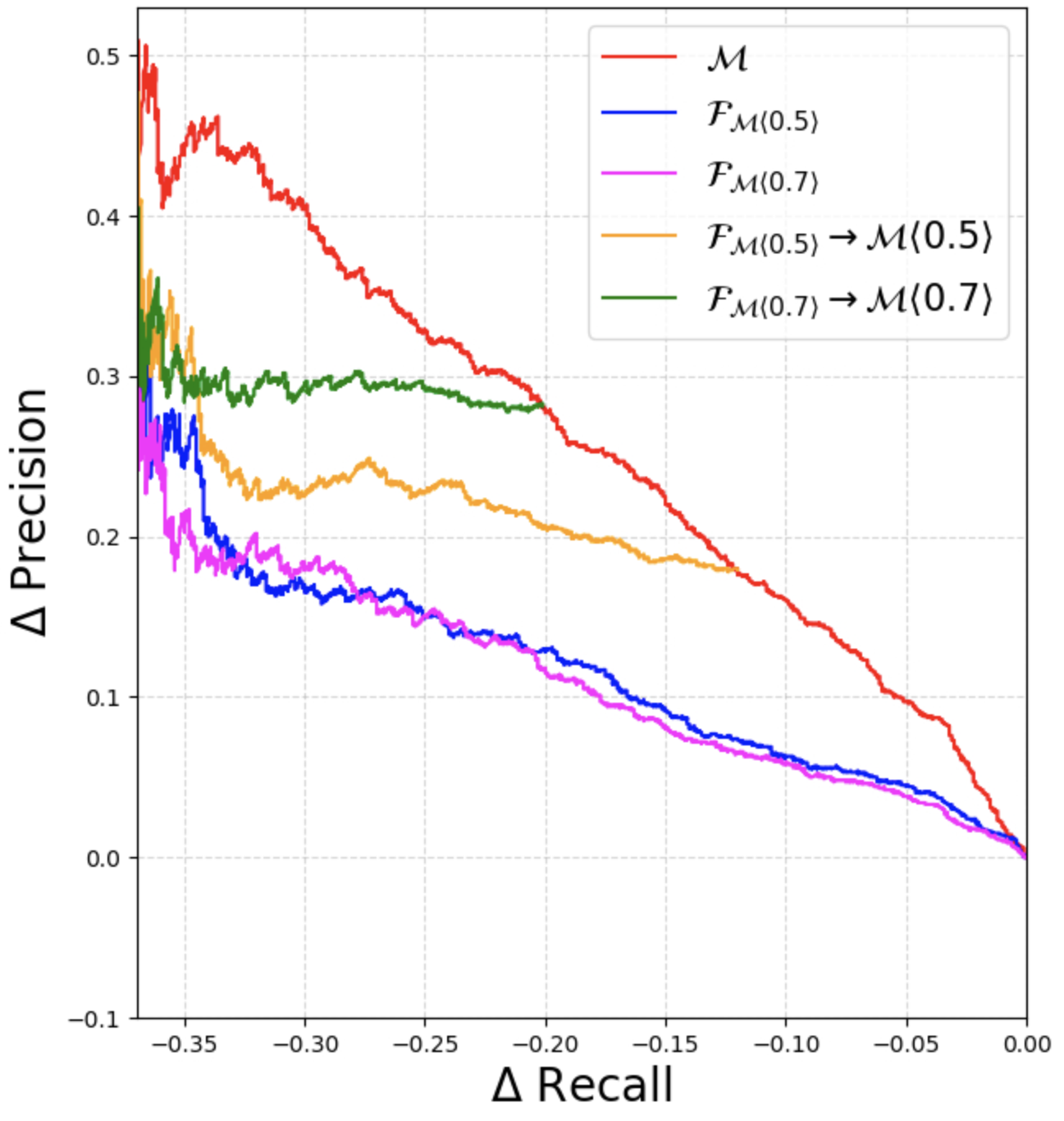}
            \caption{AQAD}
        \end{subfigure}
        \begin{subfigure}[t]{0.24\textwidth}
            \includegraphics[width=1\textwidth]{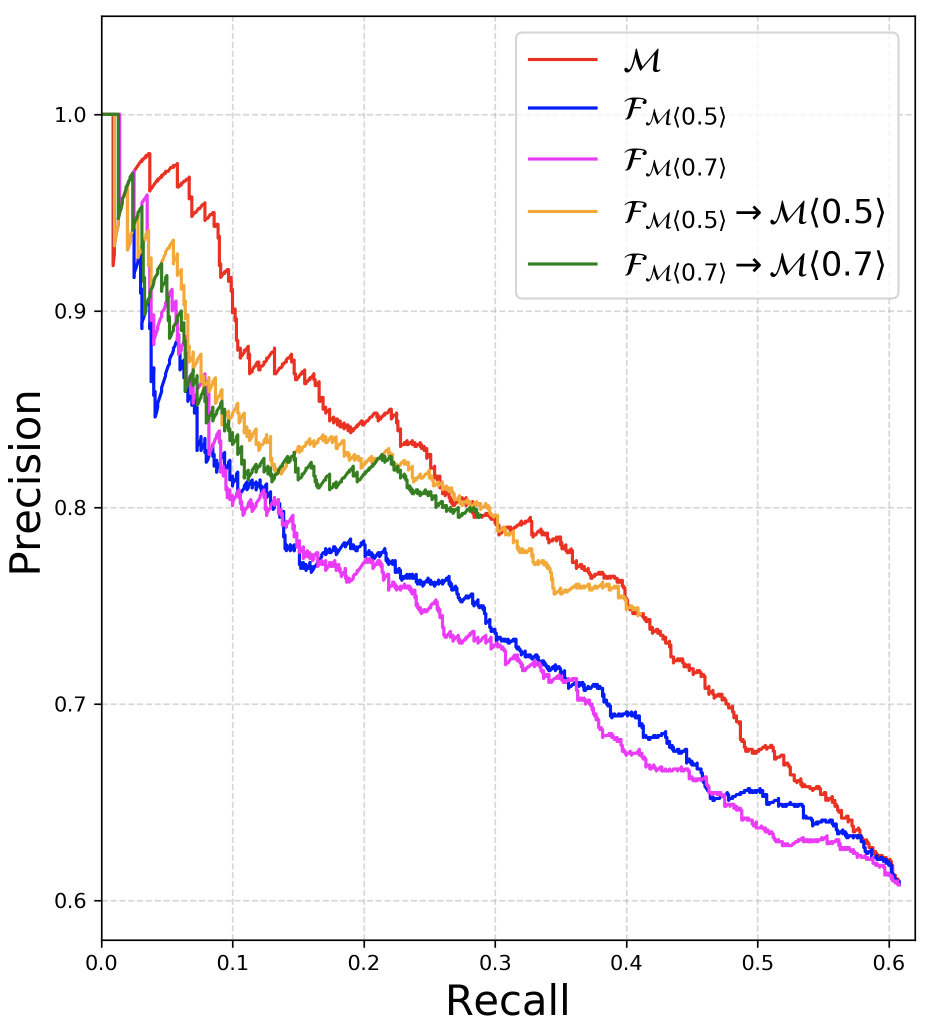}
            \caption{ASNQ}
        \end{subfigure}
        \begin{subfigure}[t]{0.24\textwidth}
            \includegraphics[width=1\textwidth]{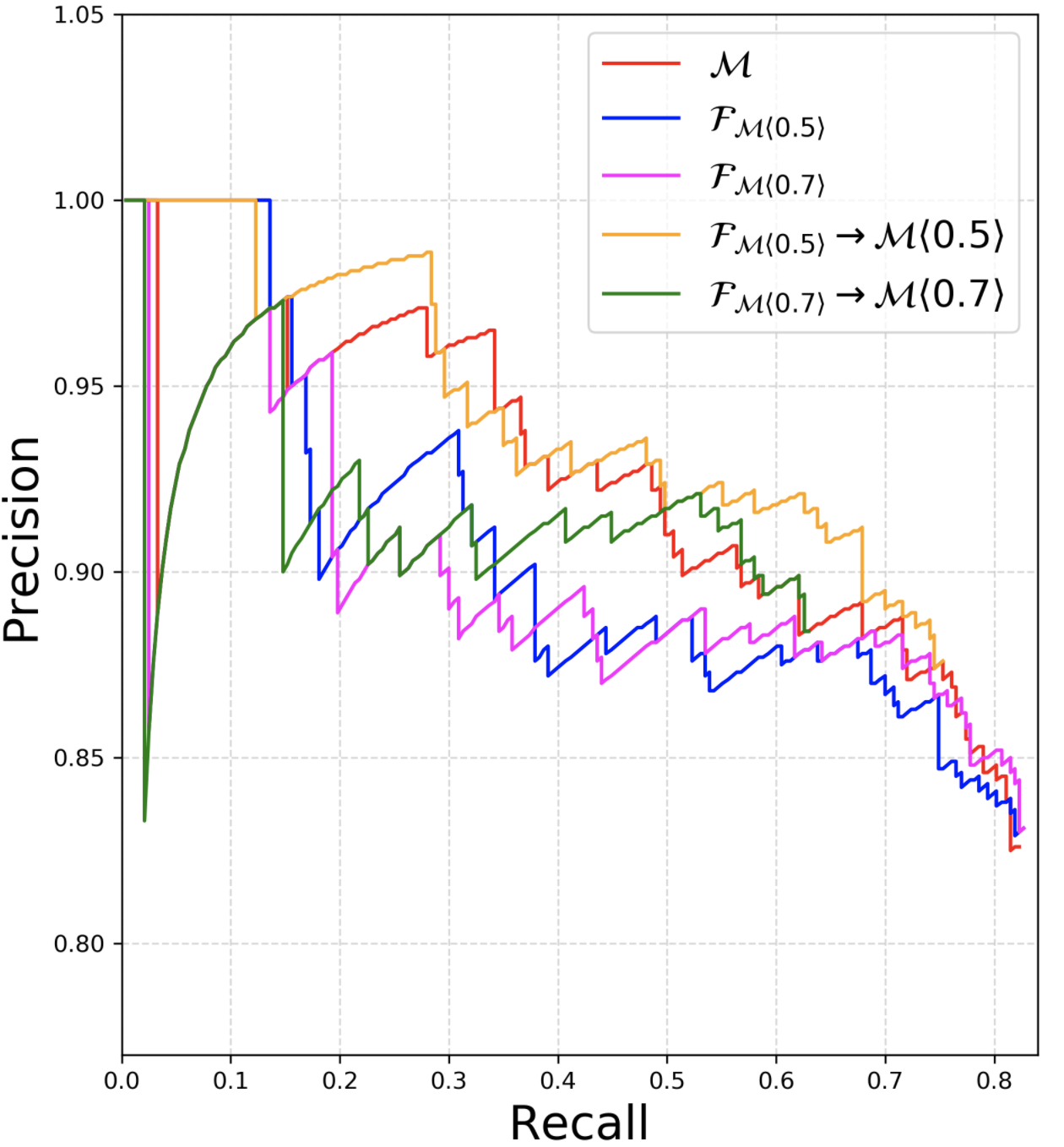}
            \caption{WikiQA}
        \end{subfigure}
    \end{subfigure}

    \caption{ Pr/Re curves for filters (\textcolor{blue}{\FMTval{0.5}} , \textcolor{magenta}{\FMTval{0.7}}), answer model \textcolor{red}{\M} and operating configurations (\textcolor{orange}{\FMTval{0.5}${\rightarrow}${\Mval{0.5}}} , \textcolor{darkgreen}{\FMTval{0.7}${\rightarrow}${\Mval{0.7}}}) are presented. For AQAD we show $\Delta$Pr/$\Delta$Re w.r.t {\Mval{0}}.}
    \label{fig:app_classifier_head_pr_re_graphs}
   
\end{figure*}

\begin{itemize}
    \item \textbf{WikiQA:} We consider the TANDA model checkpoints released by ~\citeauthor{Garg_Vu_Moschitti_2020} which are trained using sequential fine-tuning and are the state-of-the-art QA models for WikiQA. Specifically we consider the RoBERTa-Base 12-layer model first trained on ASNQ and then on WikiQA and the RoBERTa-Large-MNLI 24-layer model first trained on ASNQ and then on WikiQA. Baseline accuracy for the 12 and 24-layer model {\M} on the test split is 82.7$\%$ and 91.8$\%$ respectively.
    
    \item \textbf{ASNQ:} We consider the RoBERTa-Base and RoBERTa-Large-MNLI model checkpoints which have been fine-tuned on the training set of ASNQ for 3 epochs using a learning rate of 2e-5 Adam and have been released by ~\citeauthor{Garg_Vu_Moschitti_2020}. Baseline accuracy for the 12 and 24-layer model {\M} on the test split is 60.8$\%$ and 69.2$\%$ respectively.
    
    \begin{figure*}[t]
\centering
    \begin{subfigure}[t]{0.23\textwidth}
        \includegraphics[width=1\textwidth]{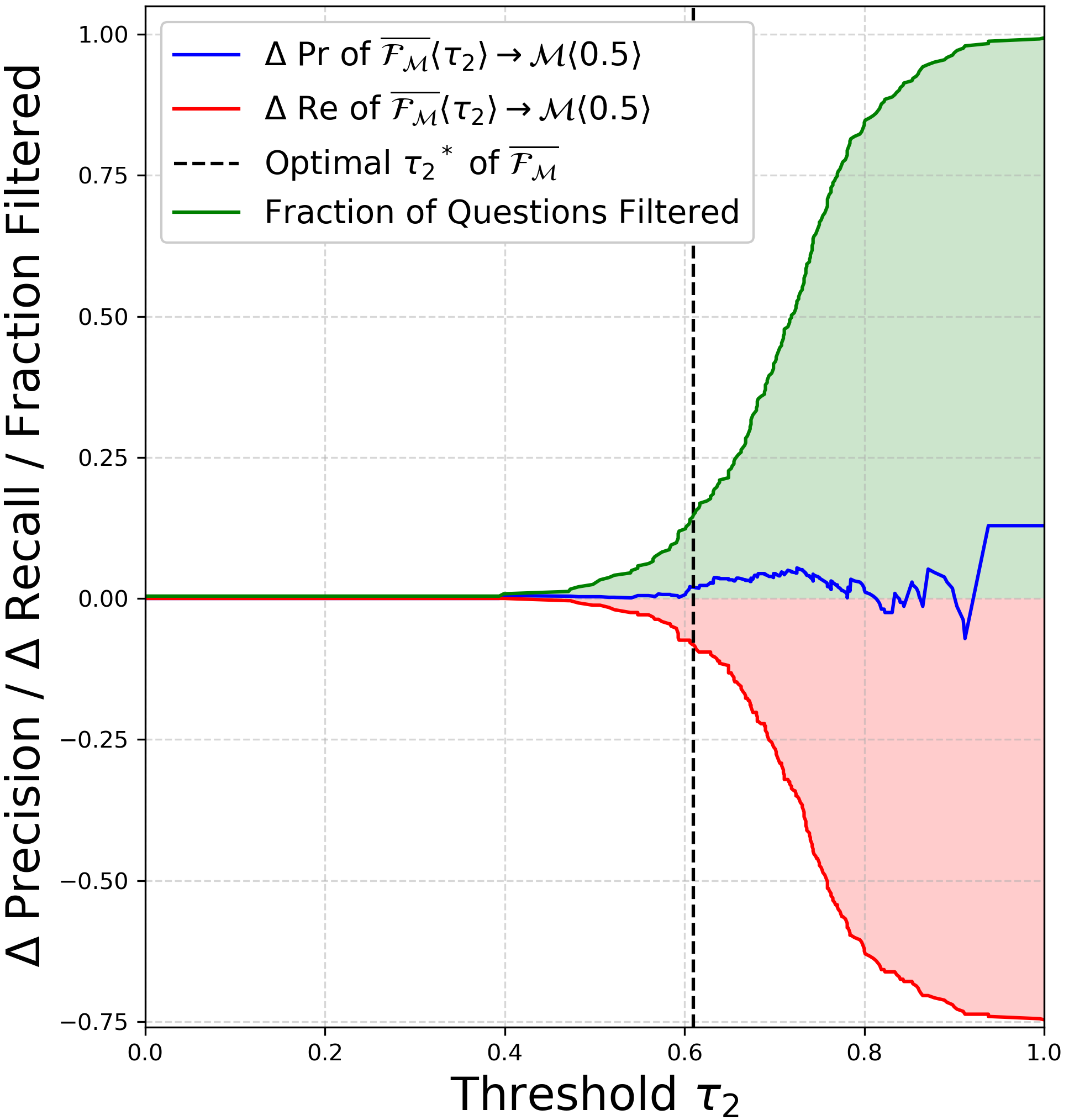}
        \caption{WikiQA, {\TM}${=}0.5$}
    \end{subfigure}
    \begin{subfigure}[t]{0.23\textwidth}
        \includegraphics[width=1\textwidth]{Plots/Experiments/Filter_Threshold_Vary/asnq_0.5.png}
        \caption{ASNQ, {\TM}${=}0.5$}
    \end{subfigure}
    \begin{subfigure}[t]{0.23\textwidth}
        \includegraphics[width=1\textwidth]{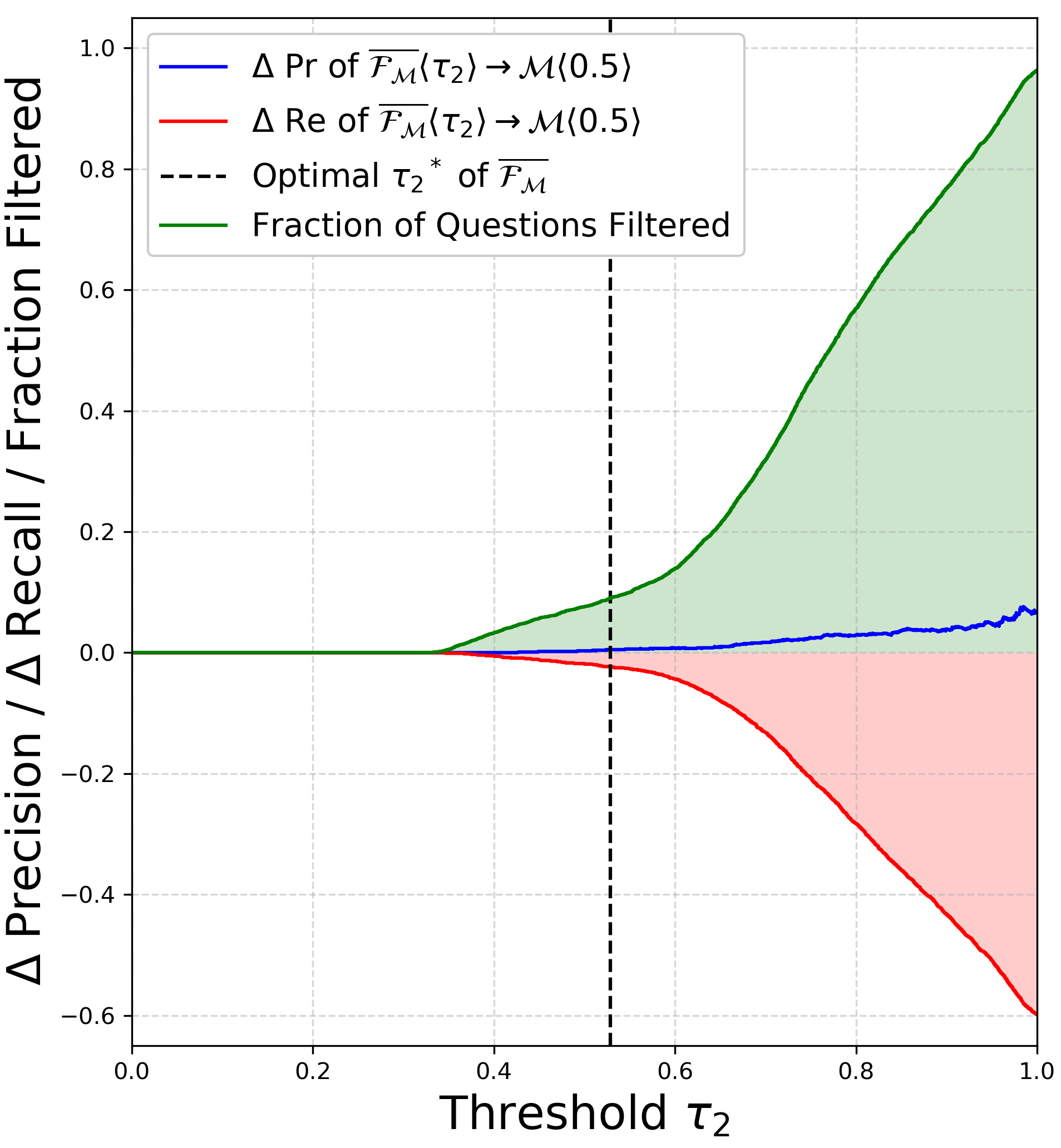}
        \caption{SQuAD 1.1, {\TM}${=}0.5$}
    \end{subfigure}
    \begin{subfigure}[t]{0.23\textwidth}
        \includegraphics[width=1\textwidth]{Plots/Experiments/Filter_Threshold_Vary/iqad_0.5.png}
        \caption{AQAD, {\TM}${=}0.5$}
    \end{subfigure}
    
    \vspace{1em}
    
    \begin{subfigure}[t]{0.23\textwidth}
        \includegraphics[width=1\textwidth]{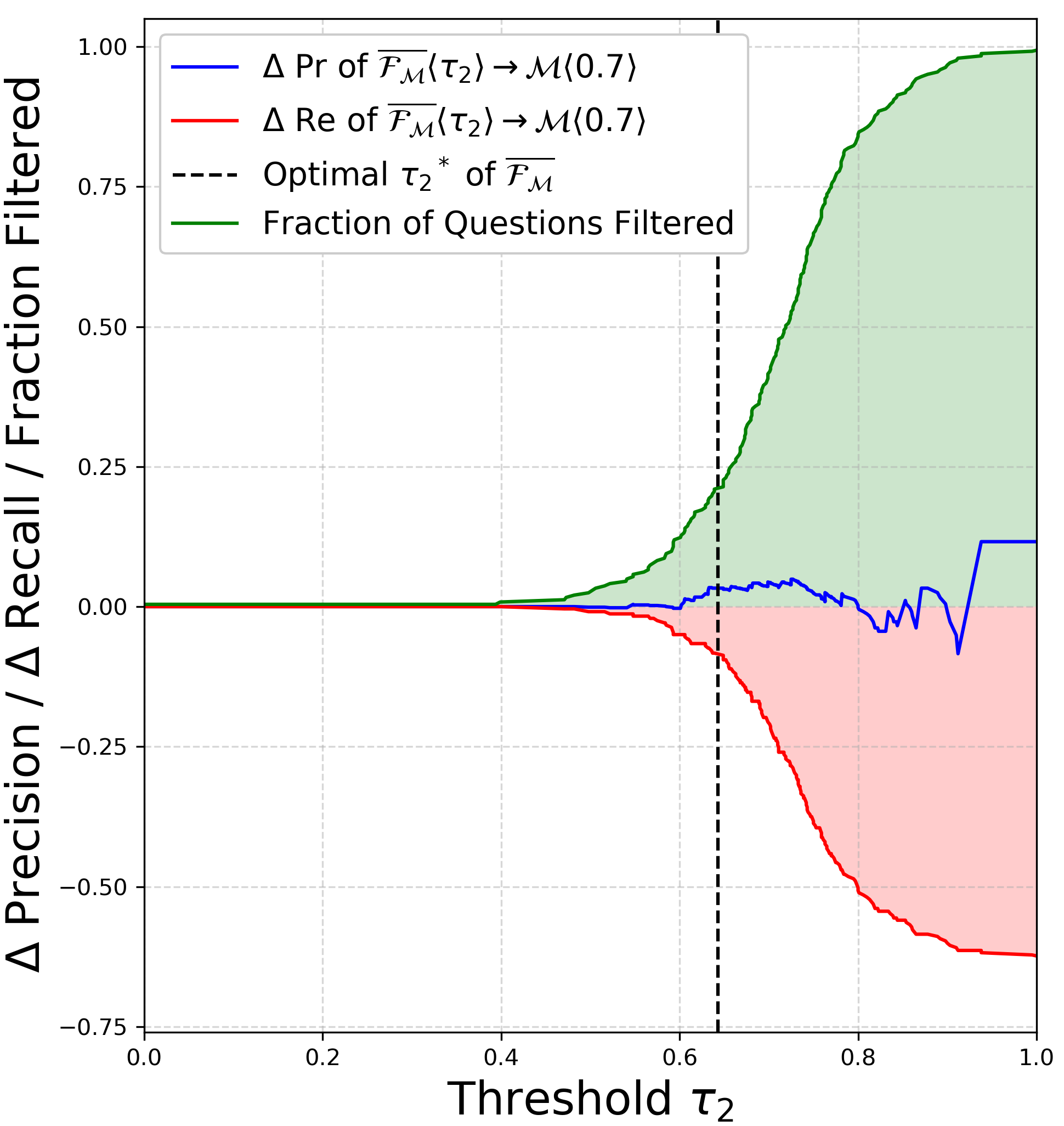}
        \caption{WikiQA, {\TM}${=}0.7$}
    \end{subfigure}
    \begin{subfigure}[t]{0.23\textwidth}
        \includegraphics[width=1\textwidth]{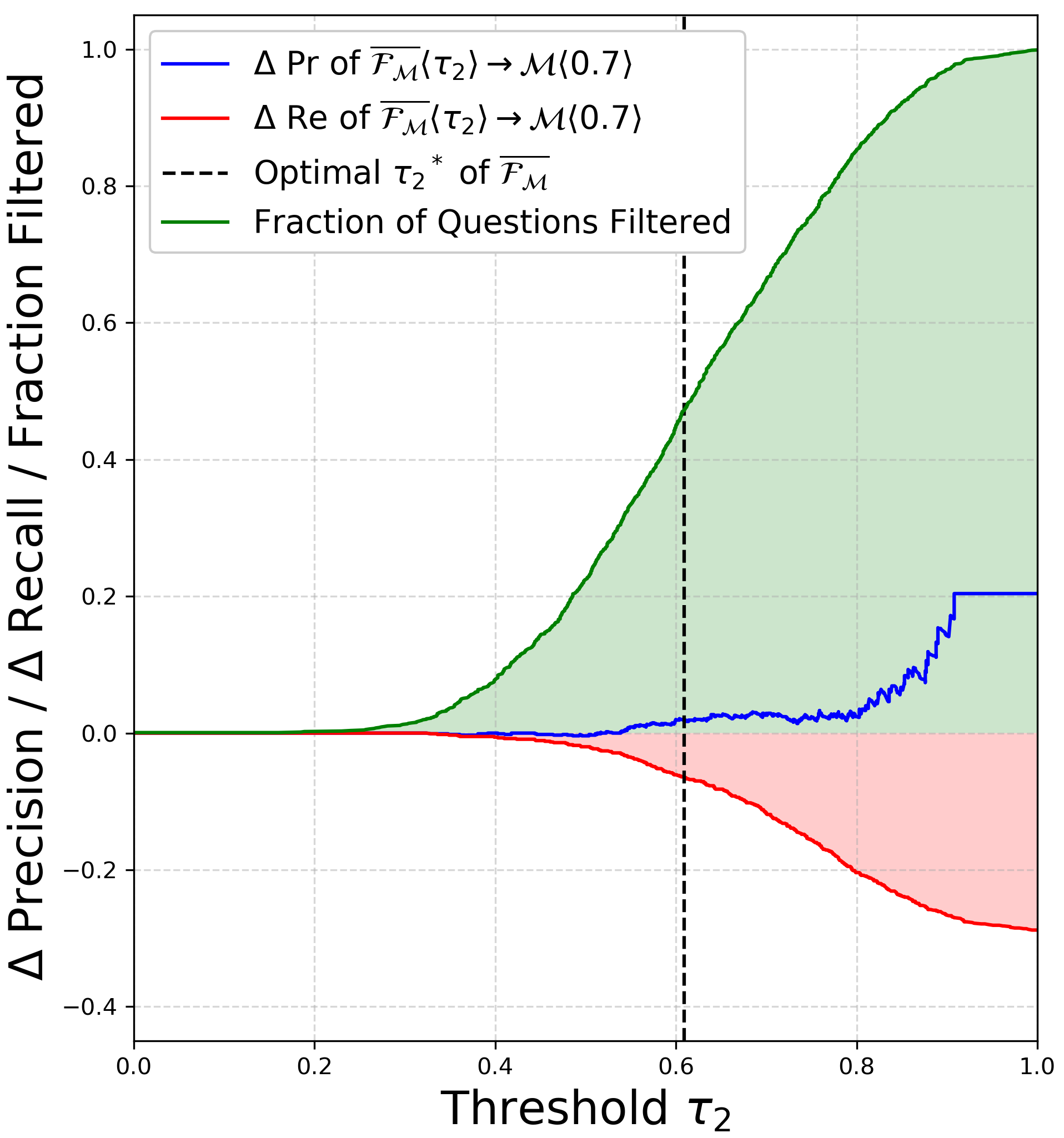}
        \caption{ASNQ, {\TM}${=}0.7$}
    \end{subfigure}
    \begin{subfigure}[t]{0.23\textwidth}
        \includegraphics[width=1\textwidth]{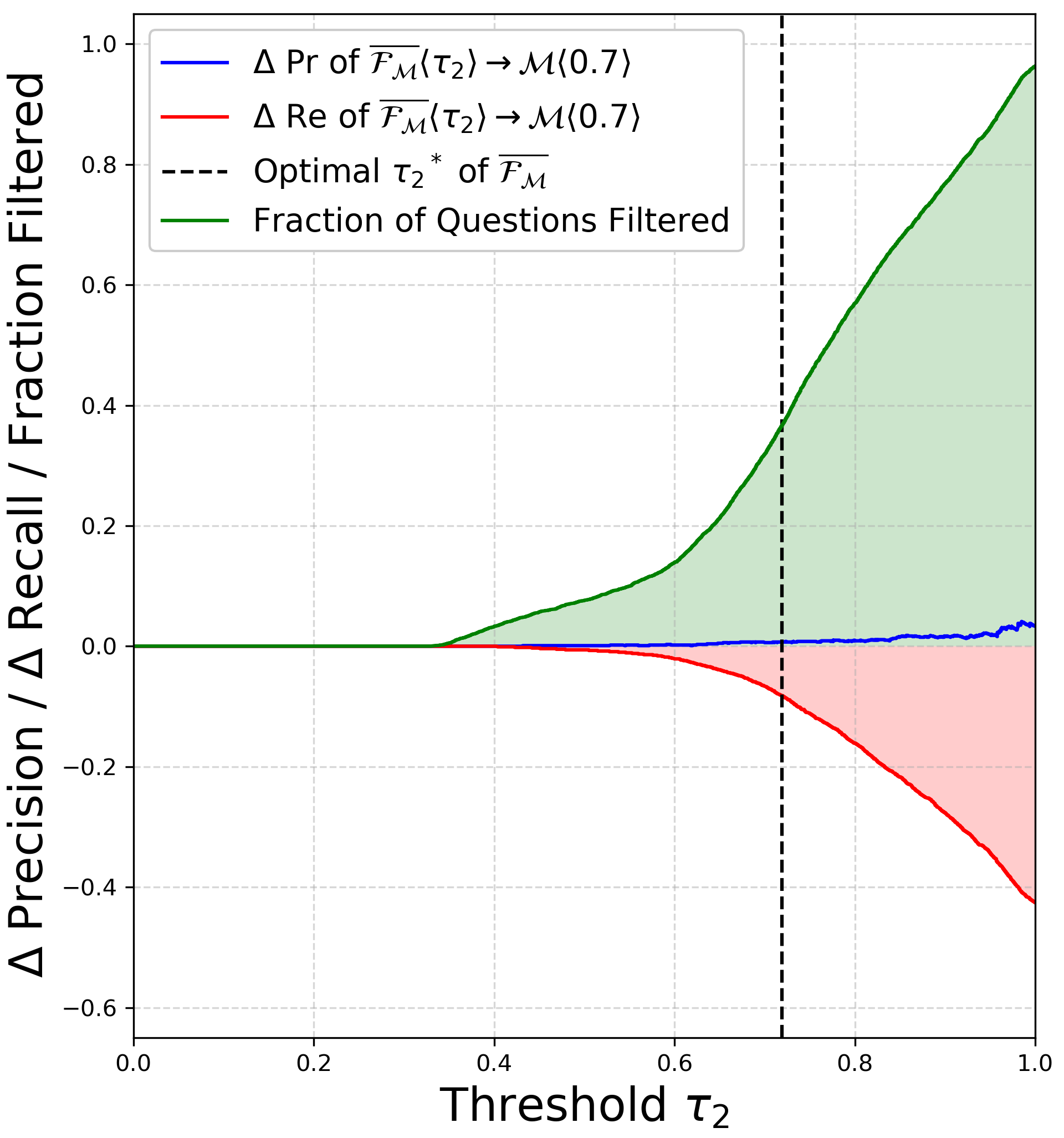}
        \caption{SQuAD 1.1, {\TM}${=}0.7$}
    \end{subfigure}
    \begin{subfigure}[t]{0.23\textwidth}
        \includegraphics[width=1\textwidth]{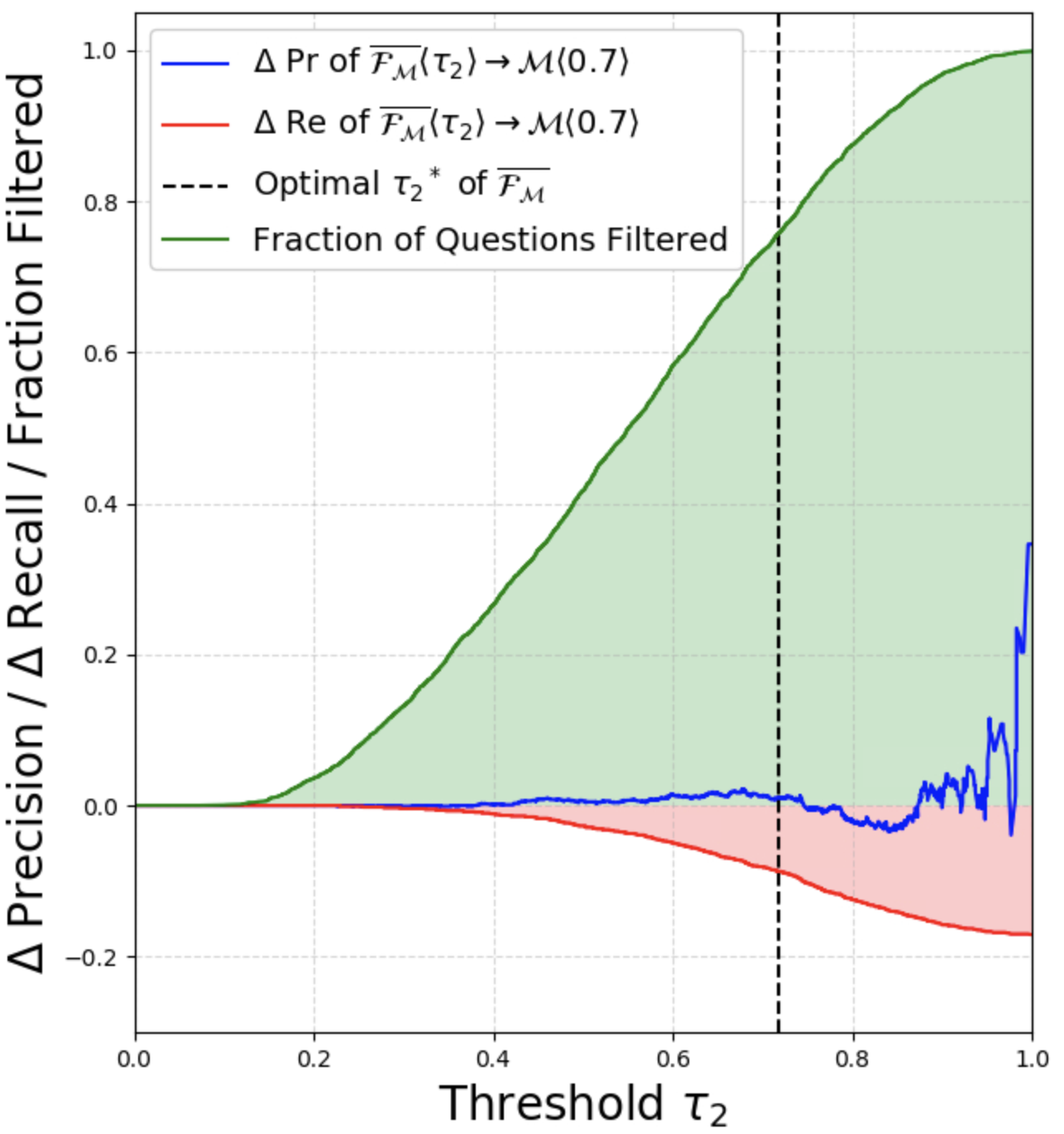}
        \caption{AQAD, {\TM}${=}0.7$}
    \end{subfigure}
    
    \vspace{1em}
    
    \begin{subfigure}[t]{0.23\textwidth}
        \includegraphics[width=1\textwidth]{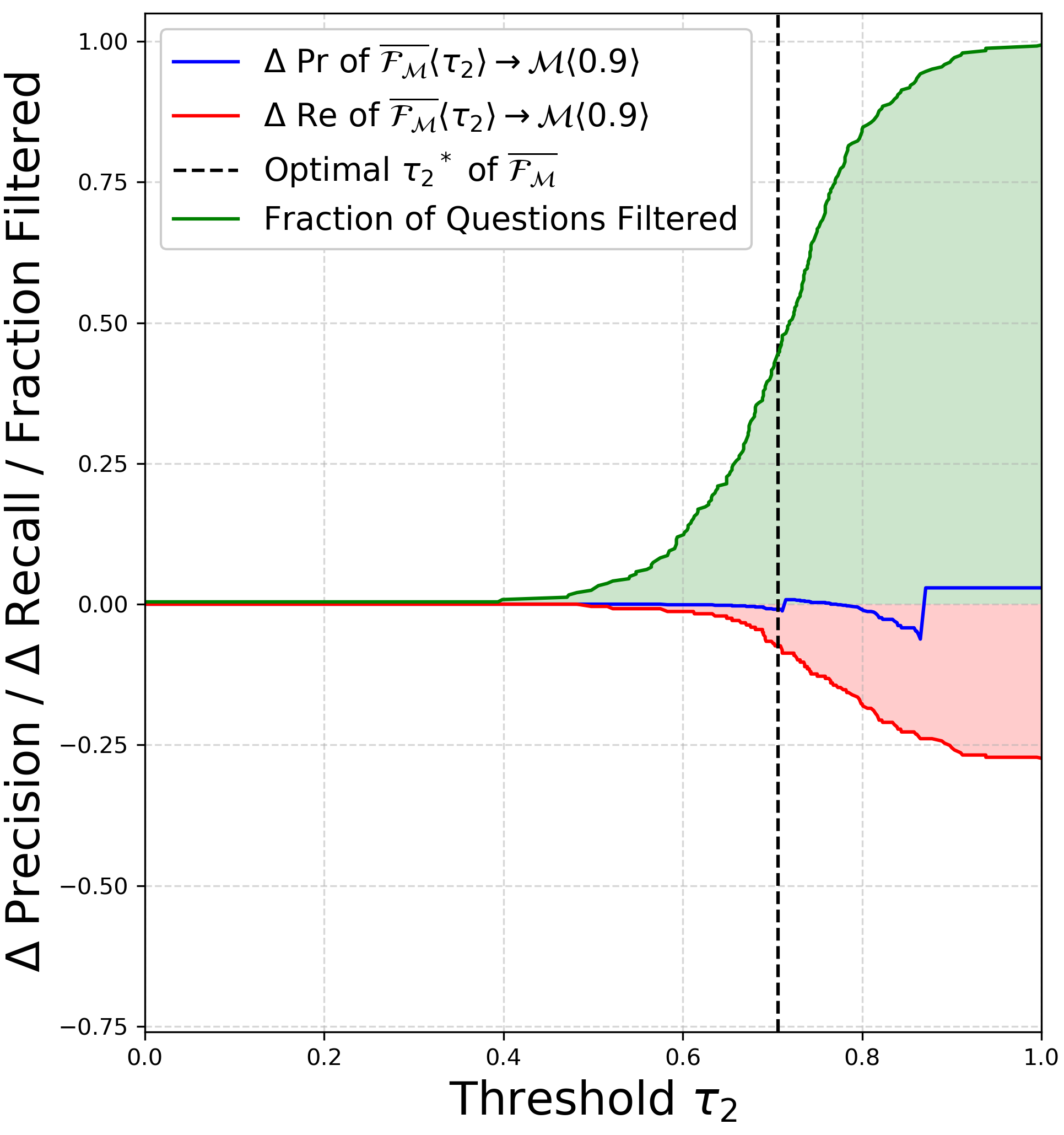}
        \caption{WikiQA, {\TM}${=}0.9$}
    \end{subfigure}
    \begin{subfigure}[t]{0.23\textwidth}
        \includegraphics[width=1\textwidth]{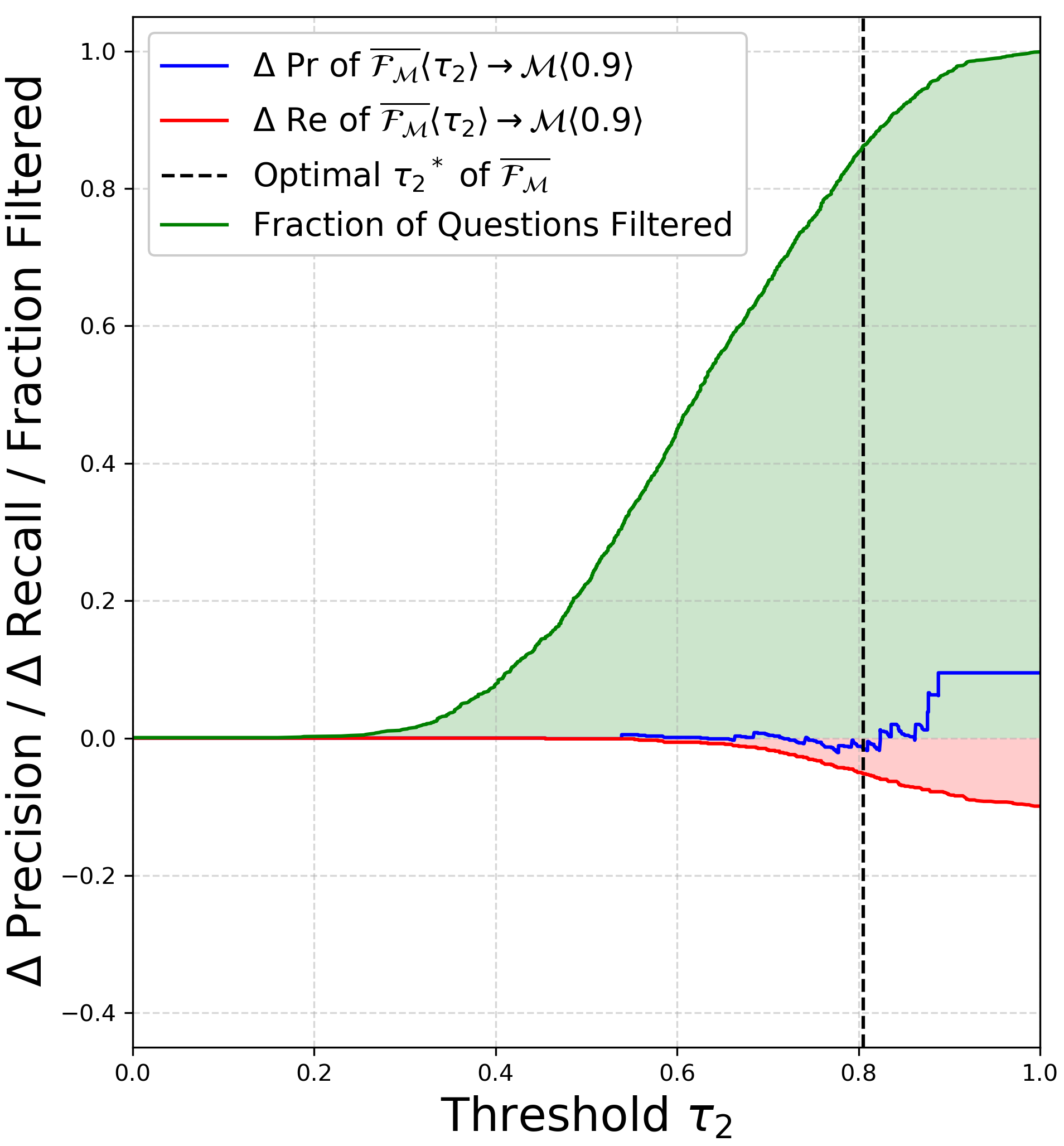}
        \caption{ASNQ, {\TM}${=}0.9$}
    \end{subfigure}
    \begin{subfigure}[t]{0.23\textwidth}
        \includegraphics[width=1\textwidth]{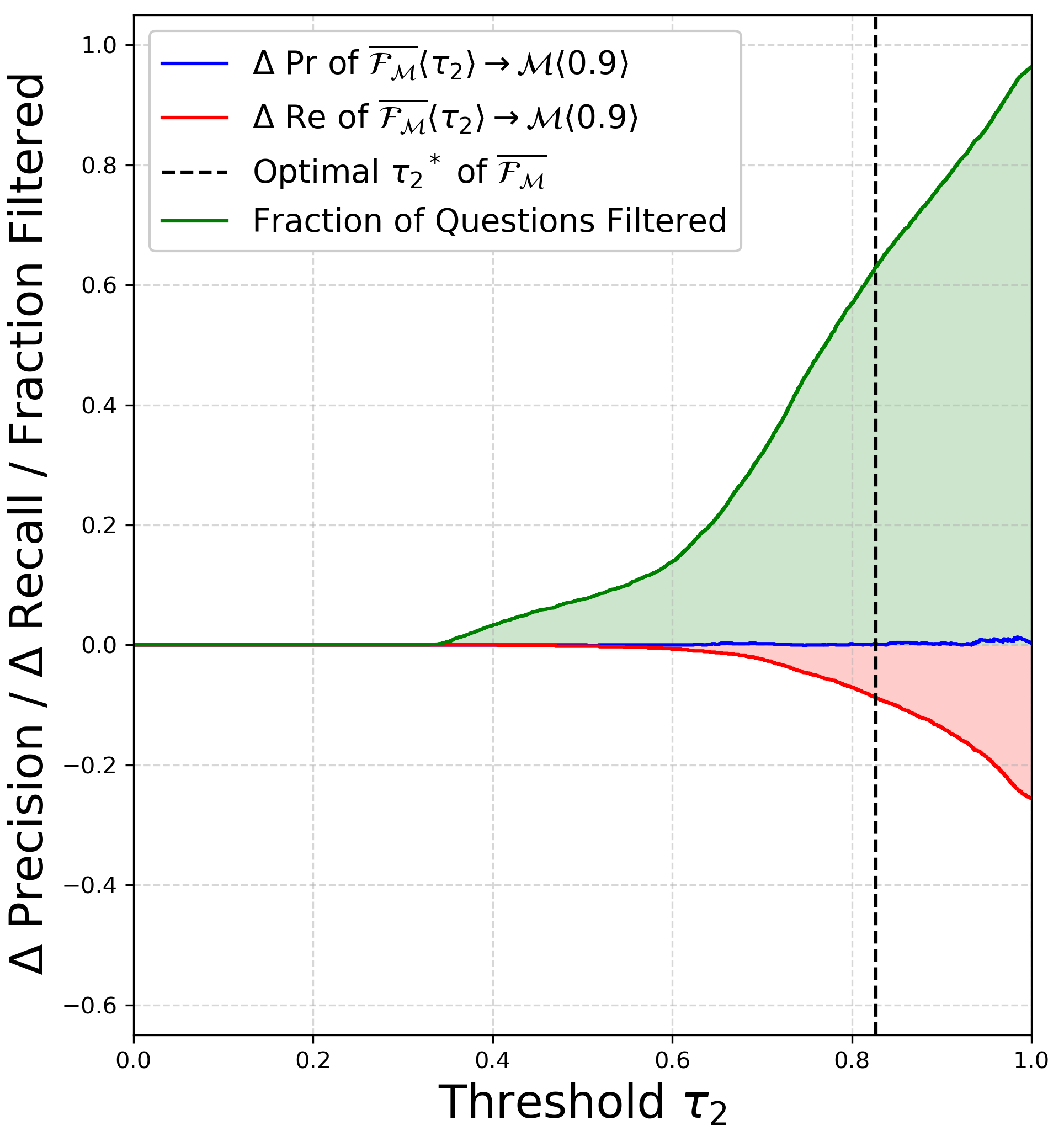}
        \caption{SQuAD 1.1, {\TM}${=}0.9$}
    \end{subfigure}
    \begin{subfigure}[t]{0.23\textwidth}
        \includegraphics[width=1\textwidth]{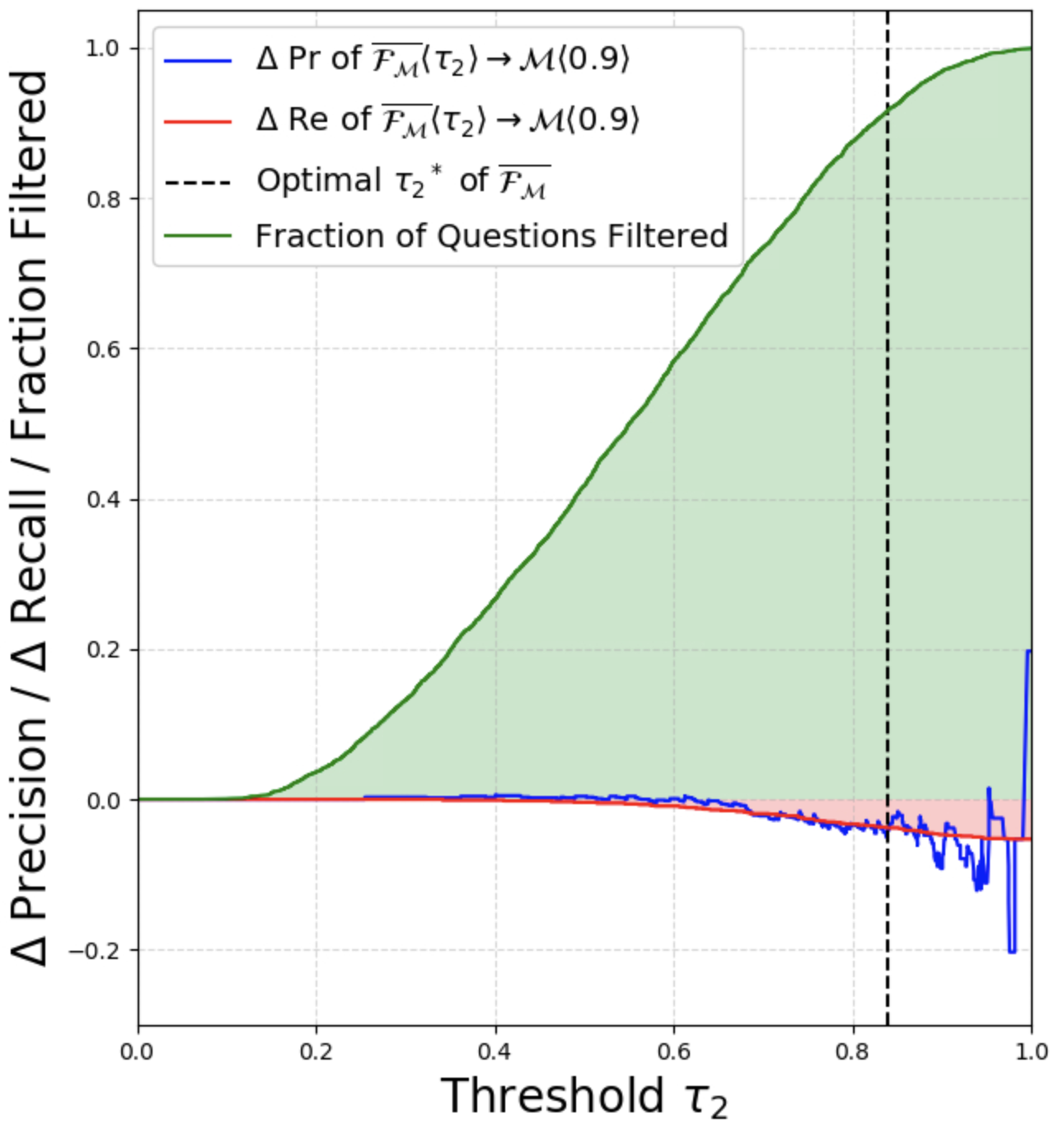}
        \caption{AQAD, {\TM}${=}0.9$}
    \end{subfigure}
   
     \vspace{5pt}
    \caption{$\Delta$ Pr/ $\Delta$ Re and fraction of questions filtered on varying {\TF} for RoBERTa-Base {\FM} on all datasets for three different {\TM} for {\M}: $\{0.5,0.7,0.9\}$ .}
    \label{fig:app_threshold_vary_plots}
\end{figure*}
    
    \item \mypara{SQuAD1.1:} We consider the BERT-Base uncased and BERT-Large uncased with whole word masking model variants and fine-tune them on the training set of SQuAD1.1 for 3 epochs with a standard learning rate of 2e-5 Adam and learning rate warm-up set for the first 5\% of training steps. Baseline accuracy for the 12 and 24-layer model {\M} on the test split is 77.9$\%$ and 84.0$\%$ respectively. 
        
    \vspace{-2pt}
    \item \mypara{AQAD:} We consider the ELECTRA-Base~\cite{clark2020electra} model and perform sequential fine-tuning using TANDA~\cite{Garg_Vu_Moschitti_2020} by a first round of fine-tuning on ASNQ for 3 epochs with a learning rate of 2e-5 Adam (learning rate warm-up of 5\%), followed by a second round of fine-tuning for 3 epochs with a learning rate of 2e-6 Adam (learning rate warm-up of 5\%). Baseline accuracy is not disclosed since the data is internal.
\end{itemize}

\section{Experimental Details}
\label{app:experimental_details}
\mypara{Training Details:} All computations are performed on NVIDIA Telsa V100 GPUs with a batch-size of 128. For training a question filter, we train {\F} using the proposed loss objectives for 3 epochs on the training split of the dataset using a standard learning rate of 2e-5 Adam (with learning rate warm-up set for the first 5$\%$ of the training steps). RoBERTa-Base and RoBERTa-Large question filters are trained corresponding to 12 and 24 layer answer models {\M} respectively. For AQAD, both the RoBERTa-Base and RoBERTa-Large question filters are trained corresponding to the ELECTRA-Base answer model {\M}. For WikiQA, the question filters are trained by sequential training: first on ASNQ for 3 epochs using a standard learning rate of 2e-5 Adam (with learning rate warm-up set for the first 5$\%$ of the training steps), and then on WikiQA for 3 epochs with a learning rate of 3e-6 (with learning rate warm-up set for first 5$\%$ of the training steps). The baseline classifier for wellformedness {\FW} and correctness of answering a question {\FC} are also trained by fine-tuning the RoBERTa-Base and Large models for 3 epochs using a standard learning rate of 2e-5 (with learning rate warm-up set for the first 5$\%$ of the training steps). As mentioned in Appendix~\ref{app:datasets}, we use an additional annotated data split containing 250k QA pairs (2.5k questions) for training {\FC} on AQAD.

\mypara{Validation Strategy:} For computing the optimal threshold {\TFBest} of a question filter {\F} as described in Section~\ref{sec:select_threshold_for_F}, we use the dev.~split of the datasets to find {\TFBest}$\in [0,1]$ such that $\mathcal{F}\langle\tau_2^*\rangle$ obtains the highest F1-score corresponding to the binary decision of answering or abstaining by {\MTM}. We present concrete results corresponding to 5 different operating thresholds {\TM} of the answer model {\M} : $\{0.3,0.5,0.6,0.7,0.9\}$. At each {\TM} for {\M}, we consider all 4 different possible question filters: our answer-model distilled question filter with regression head {\FM}, our answer-model distilled question filter with classification head {\FMTM}, the correctness question filter {\FC} and the well-formedness question filter {\FW}. For each of these four filters, we independently optimise {\TFBest}$\in [0,1]$ corresponding to best F1 filtering of {\MTM}. 

\mypara{Code:} The code for training our answer-model distilled question filters can be accessed at \url{https://github.com/alexa/wqa-question-filtering}.

\section{Complete Graphs on Approximating Pr/Re of {\M}}
\label{sec:app_pr_re_graphs}
We present Pr/Re curves of {\M} by varying $\tau_1$ (i.e, ${\mathcal{M}}{\langle\tau_1\rangle}$) and that of filters {\F}: \{{\FM},{\FC},{\FW}\} by varying $\tau_2$ (i.e, $\mathcal{F}{\langle\tau_2\rangle}$) on the dataset test splits in Fig.~\ref{fig:app_pr_re_graphs} (a)-(d). We also present Pr/Re curves comparing the filters {\F}: \{{\FM},{\FC},{\FW}\} when operating jointly with the answer model {\M} at the same threshold $\tau$, i.e, {\FM}${\langle \tau \rangle}{\rightarrow}${\MT} , {\FC}${\langle \tau \rangle}{\rightarrow}${\MT} and {\FW}${\langle \tau \rangle}{\rightarrow}${\MT}  on the dataset test splits in Fig.~\ref{fig:app_pr_re_graphs} (e)-(h). As visible from the graphs, our filter {\FM} is able to better approximate the Pr/Re of {\M} when operating independently of {\M} (Fig.~\ref{fig:app_pr_re_graphs} (a)-(d)) as well as when operating jointly with {\M} at a non-zero threshold (Fig.~\ref{fig:app_pr_re_graphs} (e)-(h)). Note that the trained answer models on WikiQA (which has a very small test set having only 237 samples) are very poorly calibrated. This is visible from the shape of the Pr/Re curve of {\M} in Fig.~\ref{fig:app_pr_re_graphs} (d),(h). Interestingly, even for such a poorly calibrated answer model, our filter {\FM} is still able to approximate the Pr/Re of {\M} better than the baselines {\FC} and {\FW}. This illustrates the validity of our technique even for very small datasets (few-shot setting).

Our classification-head filter  {\FMTM} is trained specific to a threshold {\TM} of {\M}. Since each point in the Pr/Re graph of {\M} corresponds to a different threshold {\TM}$\in [0,1]$, for fair comparison in Fig.~\ref{fig:app_pr_re_graphs}, we will need to train several {\FMTM} for every $\tau_1 \in [0,1]$ which is unfeasible. To show how the classification head filters can approximate the Pr/Re of {\M}, we arbitrarily select two thresholds {\TM}${=}\{0.5,0.7\}$ and plot the Pr/Re curves of the two classification head filters {\FMTval{0.5}} and {\FMTval{0.7}} in Fig.~\ref{fig:app_classifier_head_pr_re_graphs}. We also plot the operating configurations for these filters at the corresponding {\TM} for {\M}, i.e, {\FMTval{0.5}}${\rightarrow}${\Mval{0.5}} and {\FMTval{0.7}}${\rightarrow}${\Mval{0.7}} in Fig.~\ref{fig:app_classifier_head_pr_re_graphs}.

\section{Complete Graphs on Varying Threshold {\TF} of {\F}}
\label{sec:app_threshold_vary_plots}
We present the variation of the fraction of questions filtered by {\F} along with the change in Pr/Re of $\widehat{\Omega}$ on varying {\TF} for all the datasets in Fig.~\ref{fig:app_threshold_vary_plots}. We present plots for three different operating thresholds {\TM}${=}\{0.5,0.7,0.9\}$ of the answer model {\M}. For a dataset, since {\FM} is trained independent of any threshold {\TM}, the fraction of filtered questions would remain the same as we vary {\TF} even at different values of {\TM}. Using these graphs, one can choose the desired operating point for the filter corresponding to how much efficiency gain is desired and how much drop in recall can be tolerated.

\section{Complete Results for Optimal {\TF} of {\F}}
\label{sec:app_optimal_F_threshold}

We present the complete empirical results of our filters: {\FM} and {\FMT} at different thresholds {\TM} of {\M} in Table~\ref{tab:main_table}. We evaluate the $\%$ of questions filtered out by {\F}$\langle{\tau_2^*}\rangle$ (efficiency gains) and the resulting drop in Precision, Recall and question-answering F1 score of {\F}${\langle{\tau_2^*}\rangle}{\rightarrow}${\M}$\langle${\TM}$\rangle$ from {\MTM} on the test split of the dataset. For each dataset and model {\M}, we train one regression head question filter {\FM} and five classification head question filters {\FMTM}: one at every threshold {\TM} for {\M} $\in \{0.3,0.5,0.6,0.7,0.9\}$. The optimal filtering threshold {\TFBest} is computed using the validation strategy described in Appendix~\ref{app:experimental_details}. For the regression head {\FM}, the optimal {\TFBest} is calculated independently for every {\TM} of {\M}.

Additionally we present the complete empirical results on all datasets corresponding to the optimal filtering threshold {\TFBest} for the baseline question filters: {\FC} and {\FW} in Table~\ref{tab:app_baselines}. We observe that both {\FW} and {\FC} perform inferior in terms of filtering performance to our filters {\FM} and {\FMT}. Except for higher thresholds on AQAD, the well-formedness filter {\FW} is unable to filter out a sizable fraction of questions even when operating at {\TFBest} which indicates that human-supervised filtering of ill-formed questions is sub-optimal from an efficiency perspective. {\FC} gets better performance than {\FW}, but always trails {\FM} and {\FMT} either in terms of a smaller $\%$ of questions filtered or a larger drop in recall incurred.

\begin{table*}[t]
\resizebox{\linewidth}{!}{
\begin{tabular}{cccccccccccccc}

\multicolumn{3}{c}{\multirow{2}{*}{}}                           & \multicolumn{5}{c}{\textbf{{\M} : 12-Layer Transformer Architecture, {\F} : RoBERTa-Base}}                                                                                                             &   & \multicolumn{5}{c}{\textbf{{\M} : 24-Layer Transformer Architecture, {\F} : RoBERTa-Large}}                                                                                                                              \\ \cmidrule{4-8} \cmidrule{10-14} 
\multicolumn{3}{c}{\textbf{Threshold {\TM} for {\M} $\rightarrow$}}    & \multicolumn{1}{c}{0.3} & \multicolumn{1}{c}{0.5} & \multicolumn{1}{c}{0.6} & \multicolumn{1}{c}{0.7} & 0.9  &     & \multicolumn{1}{c}{0.3} & \multicolumn{1}{c}{0.5} & \multicolumn{1}{c}{0.6} & \multicolumn{1}{c}{0.7} & \multicolumn{1}{c}{0.9} \\ \toprule
\multirow{6}{*}{\begin{turn}{90}\textbf{WikiQA}\end{turn}} & \multirow{2}{*}{{\MTM}} & Pr / Re       	& 84.4  /  80.2 &	87.1  /  75.3 & 88.4  /  68.7 & 88.4  /  63.0 &	97.1  /  28.0 & & 92.1  /  86.0	& 92.2 / 83.1	& 92.0 / 80.2	& 92.1 / 77.0 &	95.9 / 57.2          \\ \cline{3-3}
                      &                           & F1          	& 82.2	& 80.8 & 77.3 & 	73.6	& 43.5  &   & 89.9 &	87.4 &	85.7 &	83.9 &	71.7    \\ \cmidrule{2-3}
                                          & \multirow{2}{*}{\small{\FMTM${\langle{\tau_2^*}\rangle}{\rightarrow}${\M}$\langle${\TM}$\rangle$}}         & $\Delta$ Pr / $\Delta$ Re           & \textcolor{red}{\contour{red}{-}}0.3  /  \textcolor{red}{\contour{red}{-}}2.4  & \textcolor{darkgreen}{\contour{darkgreen}{+}}0.3  / \textcolor{red}{\contour{red}{-}} 0.8  & \textcolor{red}{\contour{red}{-}}0.3  / \textcolor{red}{\contour{red}{-}}1.6  & \textcolor{darkgreen}{\contour{darkgreen}{+}}0.7  /  \textcolor{red}{\contour{red}{-}}2.5  & \textcolor{red}{\contour{red}{-}}0.4  / \textcolor{red}{\contour{red}{-}} 3.7 &  &  \textcolor{red}{\contour{red}{-}}0.1  /  \textcolor{red}{\contour{red}{-}}0.8  & \textcolor{darkgreen}{\contour{darkgreen}{+}}0.3  /  \textcolor{red}{\contour{red}{-}}2.0  & \textcolor{darkgreen}{\contour{darkgreen}{+}}0.5  / \textcolor{red}{\contour{red}{-}} 4.1  & \textcolor{red}{\contour{red}{-}}0.5  / \textcolor{red}{\contour{red}{-}} 5.0  & \textcolor{darkgreen}{\contour{darkgreen}{+}}0.2 / \textcolor{red}{\contour{red}{-}}6.1      \\ \cline{3-3}
                      &                           & $\%$ Filter / $\Delta$ F1 & 	4.1 / \textcolor{red}{\contour{red}{-}}0.9 &	2.9 / \textcolor{red}{\contour{red}{-}}0.4 & 3.7 / \textcolor{red}{\contour{red}{-}}1.1	&7.0 / \textcolor{red}{\contour{red}{-}}1.5 &	42.8 / \textcolor{red}{\contour{red}{-}} 4.7 &   &  0.8 / \textcolor{red}{\contour{red}{-}}0.4  & 3.3 / \textcolor{red}{\contour{red}{-}} 1.0  & 6.6 / \textcolor{red}{\contour{red}{-}} 2.2  & 7.8 / \textcolor{red}{\contour{red}{-}} 3.3  & 18.9 / \textcolor{red}{\contour{red}{-}} 4.1 \\ \cmidrule{2-3}
                     & \multirow{2}{*}{\small{\FM${\langle{\tau_2^*}\rangle}{\rightarrow}${\M}$\langle${\TM}$\rangle$}}         & $\Delta$ Pr / $\Delta$ Re   & \textcolor{darkgreen}{\contour{darkgreen}{+}}0.7  / \textcolor{red}{\contour{red}{-}}2.8  & \textcolor{darkgreen}{\contour{darkgreen}{+}}2.3  /  \textcolor{red}{\contour{red}{-}}10.3  & \textcolor{darkgreen}{\contour{darkgreen}{+}}2.8  / \textcolor{red}{\contour{red}{-}} 9.0  & \textcolor{darkgreen}{\contour{darkgreen}{+}}3.3  / \textcolor{red}{\contour{red}{-}}8.3  & \textcolor{red}{\contour{red}{-}}0.9  / \textcolor{red}{\contour{red}{-}}7.4 &   &  \textcolor{red}{\contour{red}{-}}0.1  / \textcolor{red}{\contour{red}{-}}0.8  & 0  / \textcolor{red}{\contour{red}{-}}0.8  & \textcolor{red}{\contour{red}{-}}0.1  / \textcolor{red}{\contour{red}{-}}0.4  & \textcolor{red}{\contour{red}{-}}0.2  / \textcolor{red}{\contour{red}{-}}2.1  & \textcolor{red}{\contour{red}{-}}0.2  / \textcolor{red}{\contour{red}{-}}2.9  \\ \cline{3-3}
                      &                           & $\%$ Filter / $\Delta$ F1 &  4.1 / \textcolor{red}{\contour{red}{-}} 1.1 &	17.3 / \textcolor{red}{\contour{red}{-}} 5.0 & 20.9 / \textcolor{red}{\contour{red}{-}} 5.1 &	21.0 / \textcolor{red}{\contour{red}{-}}5.1 &	44.0 / \textcolor{red}{\contour{red}{-}} 9.6  &    &  1.2 / \textcolor{red}{\contour{red}{-}}1.4  & 1.8 / \textcolor{red}{\contour{red}{-}}0.4  & 2.6 / \textcolor{red}{\contour{red}{-}}0.3  & 3.9 / \textcolor{red}{\contour{red}{-}}1.4  & 8.8 / \textcolor{red}{\contour{red}{-}}2.4                  \\ \midrule
\multirow{6}{*}{\begin{turn}{90}\textbf{ASNQ}\end{turn}} & \multirow{2}{*}{{\MTM}} & Pr / Re                & 68.2  /  48.7                & 74.6  /  41.1                &           77.2  /  36.1              & 79.6  /  28.9      & 90.5  /  10.0 &   & 75.7  /  61.1               & 79.6  /  54.4    &         81.9  /  49.3            & 84.5  /  42.7               & 92.9  /  20.7     \\ \cline{3-3}
                      &                           & F1                               & 56.8                     & 53.0                     &                 49.2         & 42.4                     & 18.0   &           & 67.6                     & 64.6                     &         61.5        & 56.7                    & 33.9      \\ \cmidrule{2-3}
                                          & \multirow{2}{*}{\small{\FMTM${\langle{\tau_2^*}\rangle}{\rightarrow}${\M}$\langle${\TM}$\rangle$}}         & $\Delta$ Pr / $\Delta$ Re         & \textcolor{darkgreen}{\contour{darkgreen}{+}}0.9  / \textcolor{red}{\contour{red}{-}}1.8                & \textcolor{darkgreen}{\contour{darkgreen}{+}}1.6  / \textcolor{red}{\contour{red}{-}}2.9               &     \textcolor{darkgreen}{\contour{darkgreen}{+}}1.6  /  \textcolor{red}{\contour{red}{-}}4.7  & \textcolor{darkgreen}{\contour{darkgreen}{+}}2.5 / \textcolor{red}{\contour{red}{-}}8.2                & \textcolor{darkgreen}{\contour{darkgreen}{+}}0.6  /  \textcolor{red}{\contour{red}{-}}3.9  &  & 0  /  0                              & \textcolor{darkgreen}{\contour{darkgreen}{+}}1.3  / \textcolor{red}{\contour{red}{-}} 3.2  &     \textcolor{darkgreen}{\contour{darkgreen}{+}}1.5  / \textcolor{red}{\contour{red}{-}}3.0           & \textcolor{darkgreen}{\contour{darkgreen}{+}}1.8  / \textcolor{red}{\contour{red}{-}}7.4               & \textcolor{darkgreen}{\contour{darkgreen}{+}}2.4 / \textcolor{red}{\contour{red}{-}}7.0                \\ \cline{3-3}
                      &                           & $\%$ Filter / $\Delta$ F1 &  7.8 / \textcolor{red}{\contour{red}{-}}0.9                      & 17.8 / \textcolor{red}{\contour{red}{-}}2.1                     &                 29.8 / \textcolor{red}{\contour{red}{-}}4.3         & 54.2 / \textcolor{red}{\contour{red}{-}}9.3                     & 83.8 / \textcolor{red}{\contour{red}{-}}6.6   &   &  0.2 / 0                      & 10.5 / \textcolor{red}{\contour{red}{-}}1.9                     &   15.6 / \textcolor{red}{\contour{red}{-}}2.0              & 29.8 / \textcolor{red}{\contour{red}{-}}6.6                     & 66.2 / \textcolor{red}{\contour{red}{-}}9.9                     \\ \cmidrule{2-3}
           & \multirow{2}{*}{\small{\FM${\langle{\tau_2^*}\rangle}{\rightarrow}${\M}$\langle${\TM}$\rangle$}}         & $\Delta$ Pr / $\Delta$ Re         &  \textcolor{darkgreen}{\contour{darkgreen}{+}}0.7 /  \textcolor{red}{\contour{red}{-}}1.2  & \textcolor{darkgreen}{\contour{darkgreen}{+}}0.6  /  \textcolor{red}{\contour{red}{-}}2.7  & \textcolor{darkgreen}{\contour{darkgreen}{+}}1.4  /  \textcolor{red}{\contour{red}{-}}5.3  & \textcolor{darkgreen}{\contour{darkgreen}{+}}1.9  / \textcolor{red}{\contour{red}{-}}6.4  & \textcolor{red}{\contour{red}{-}}1.3 / \textcolor{red}{\contour{red}{-}}5.1 &  &  \textcolor{darkgreen}{\contour{darkgreen}{+}}0.2  /  \textcolor{red}{\contour{red}{-}}0.1  & \textcolor{darkgreen}{\contour{darkgreen}{+}}0.8  /  \textcolor{red}{\contour{red}{-}}3.3  & \textcolor{darkgreen}{\contour{darkgreen}{+}}1.0  /  \textcolor{red}{\contour{red}{-}}3.5  & \textcolor{darkgreen}{\contour{darkgreen}{+}}1.4  / \textcolor{red}{\contour{red}{-}}3.9  & \textcolor{darkgreen}{\contour{darkgreen}{+}}2.2  /  \textcolor{red}{\contour{red}{-}}6.3                \\ \cline{3-3}
                      &                           & $\%$ Filter / $\Delta$ F1 &  6.0 / \textcolor{red}{\contour{red}{-}}0.4                      & 14.9 / \textcolor{red}{\contour{red}{-}}2.2                     &       33.5 / \textcolor{red}{\contour{red}{-}}4.9                   & 47.1 / \textcolor{red}{\contour{red}{-}}7.1                     & 86.0 / \textcolor{red}{\contour{red}{-}}8.7  &    &  1.0 / 0                      & 12.1 / \textcolor{red}{\contour{red}{-}}2.1                     &         16.1 / \textcolor{red}{\contour{red}{-}}2.5                 & 21.6 / \textcolor{red}{\contour{red}{-}}3.2                     & 61.6 / \textcolor{red}{\contour{red}{-}}8.9                     \\ \midrule

\multirow{6}{*}{\begin{turn}{90}\textbf{SQuAD 1.1}\end{turn}} & \multirow{2}{*}{{\MTM}} & Pr / Re         &     	82.0  /  75.0  & 	88.7 / 63.3 & 	91.0  /  54.6 & 93.4  /  45.9 & 	96.7  /  28.7 &  & 	86.0 / 82.9 & 	90.1 / 75.3 & 	92.3 / 67.5 & 	94.4 / 59.7 & 	97.6 / 42.4         \\ \cline{3-3}
                      &                           & F1             & 78.3 & 	73.9 &	68.3 &	61.6 & 44.3   &   &     	84.4 & 	82.0 & 	78.0 & 	73.1 & 	59.1          \\ \cmidrule{2-3}
            & \multirow{2}{*}{\small{\FMTM${\langle{\tau_2^*}\rangle}{\rightarrow}${\M}$\langle${\TM}$\rangle$}}         & $\Delta$ Pr / $\Delta$ Re         & 0 / 0  & \textcolor{darkgreen}{\contour{darkgreen}{+}}0.3  / \textcolor{red}{\contour{red}{-}} 2.3  & \textcolor{darkgreen}{\contour{darkgreen}{+}}0.4  / \textcolor{red}{\contour{red}{-}}2.4  & \textcolor{darkgreen}{\contour{darkgreen}{+}}0.6  / \textcolor{red}{\contour{red}{-}}4.8  & \textcolor{darkgreen}{\contour{darkgreen}{+}}0.1  /  \textcolor{red}{\contour{red}{-}}7.7  &   &  0 / 0  & \textcolor{darkgreen}{\contour{darkgreen}{+}}0.1  / \textcolor{red}{\contour{red}{-}}0.5  & \textcolor{darkgreen}{\contour{darkgreen}{+}}0.3  / \textcolor{red}{\contour{red}{-}}1.8  & \textcolor{darkgreen}{\contour{darkgreen}{+}}0.5 / \textcolor{red}{\contour{red}{-}}3.8 & 	\textcolor{darkgreen}{\contour{darkgreen}{+}}0.3 / \textcolor{red}{\contour{red}{-}}3.7               \\ \cline{3-3}
                      &                           & $\%$ Filter / $\Delta$ F1  &	0 / 0	 & 9.4 / \textcolor{red}{\contour{red}{-}}1.5 &	12.9 / \textcolor{red}{\contour{red}{-}}1.9 &	27.4 / \textcolor{red}{\contour{red}{-}}4.4 &	61.0 / \textcolor{red}{\contour{red}{-}}9.8 &  &	0 / 0 &	2.0 / \textcolor{red}{\contour{red}{-}}0.2 &	6.4 / \textcolor{red}{\contour{red}{-}}1.1	&14.0 / \textcolor{red}{\contour{red}{-}}2.7 &	47.5 / \textcolor{red}{\contour{red}{-}}3.7 \\ \cmidrule{2-3}
            & \multirow{2}{*}{\small{\FM${\langle{\tau_2^*}\rangle}{\rightarrow}${\M}$\langle${\TM}$\rangle$}}         & $\Delta$ Pr / $\Delta$ Re           & \textcolor{darkgreen}{\contour{darkgreen}{+}}0.2  / \textcolor{red}{\contour{red}{-}}0.4  & \textcolor{darkgreen}{\contour{darkgreen}{+}}0.5  /  \textcolor{red}{\contour{red}{-}}2.3  & \textcolor{darkgreen}{\contour{darkgreen}{+}}0.6  /  \textcolor{red}{\contour{red}{-}}3.2  & \textcolor{darkgreen}{\contour{darkgreen}{+}}0.7  /  \textcolor{red}{\contour{red}{-}}8.2  & \textcolor{darkgreen}{\contour{darkgreen}{+}}0.2  / \textcolor{red}{\contour{red}{-}}8.8 &      & 0  /  0 & \textcolor{darkgreen}{\contour{darkgreen}{+}}0.1  / \textcolor{red}{\contour{red}{-}}0.5  & \textcolor{darkgreen}{\contour{darkgreen}{+}}0.3 / \textcolor{red}{\contour{red}{-}}1.2 & \textcolor{darkgreen}{\contour{darkgreen}{+}}0.7  / \textcolor{red}{\contour{red}{-}}5.5  & \textcolor{darkgreen}{\contour{darkgreen}{+}}0.3  /  \textcolor{red}{\contour{red}{-}}8.1  \\ \cline{3-3}
                      &                           & $\%$ Filter / $\Delta$ F1  &	1.1 / \textcolor{red}{\contour{red}{-}}0.1 &	9.0 / \textcolor{red}{\contour{red}{-}}1.4 & 	14.2 / \textcolor{red}{\contour{red}{-}}2.5 & 	36.5 / \textcolor{red}{\contour{red}{-}}7.8 & 	63.0 / \textcolor{red}{\contour{red}{-}}10.3  & &    	0 / 0 &	2.4 / \textcolor{red}{\contour{red}{-}}0.3 &	5.1 / \textcolor{red}{\contour{red}{-}}0.7 &	18.2 / \textcolor{red}{\contour{red}{-}}4.1 & 41.8 / \textcolor{red}{\contour{red}{-}}8.3                \\ \midrule 
\multirow{5}{*}{\begin{turn}{90}\textbf{AQAD} \end{turn}} & {\MTM} & Pr / Re         &  	{$\uparrow$}9.2 / {$\downarrow$}4.5 & 	{$\uparrow$}17.9 / {$\downarrow$}12.1 & 	{$\uparrow$}23.4 / {$\downarrow$}15.7 & 	{$\uparrow$}28.1 / {$\downarrow$}20.1 & 	{$\uparrow$}43.0 / {$\downarrow$}31.9 &  &  	{$\uparrow$}9.2 / {$\downarrow$}4.5 & 	{$\uparrow$}17.9 / {$\downarrow$}12.1 & 	{$\uparrow$}23.4 / {$\downarrow$}15.7 & 	{$\uparrow$}28.1 / {$\downarrow$}20.1 & 	{$\uparrow$}43.0 / {$\downarrow$}31.9     \\ 
             \cmidrule{2-3}
     & \multirow{2}{*}{\small{\FMTM${\langle{\tau_2^*}\rangle}{\rightarrow}${\M}$\langle${\TM}$\rangle$}}         & $\Delta$ Pr / $\Delta$ Re           & \textcolor{darkgreen}{\contour{darkgreen}{+}}1.3  /  \textcolor{red}{\contour{red}{-}}3.4  & \textcolor{darkgreen}{\contour{darkgreen}{+}}1.9  /  \textcolor{red}{\contour{red}{-}}5.0  & \textcolor{darkgreen}{\contour{darkgreen}{+}}2.0  /  \textcolor{red}{\contour{red}{-}}5.6  & \textcolor{darkgreen}{\contour{darkgreen}{+}}1.2  /  \textcolor{red}{\contour{red}{-}}6.0  & \textcolor{red}{\contour{red}{-}}1.9  /  \textcolor{red}{\contour{red}{-}}3.4 & 
   &  \textcolor{darkgreen}{\contour{darkgreen}{+}}1.8  /  \textcolor{red}{\contour{red}{-}}3.2  &\textcolor{darkgreen}{\contour{darkgreen}{+}}2.5  /  \textcolor{red}{\contour{red}{-}}4.9  & \textcolor{darkgreen}{\contour{darkgreen}{+}}2.3  /  \textcolor{red}{\contour{red}{-}}5.5  & \textcolor{darkgreen}{\contour{darkgreen}{+}}1.8  /  \textcolor{red}{\contour{red}{-}}4.6  & \textcolor{darkgreen}{\contour{darkgreen}{+}}1.9  /  \textcolor{red}{\contour{red}{-}}3.0  \\ \cline{3-3}
                \cline{3-3}
              &                & $\%$ Filter / $\Delta$ F1 &  20.8 / \textcolor{red}{\contour{red}{-}}2.1 & 43.2 / \textcolor{red}{\contour{red}{-}}4.8 &	53.9 / \textcolor{red}{\contour{red}{-}}6.4 &	65.2 / \textcolor{red}{\contour{red}{-}}8.0 &	89.4 / \textcolor{red}{\contour{red}{-}}6.2 &  & 	21.9 / \textcolor{red}{\contour{red}{-}}1.8 &	45.8 / \textcolor{red}{\contour{red}{-}}4.6 &	56.9 / \textcolor{red}{\contour{red}{-}}6.3 &	66.2 / \textcolor{red}{\contour{red}{-}}7.4 &	89.9 / \textcolor{red}{\contour{red}{-}}5.4
 \\ \cmidrule{2-3}
                 & \multirow{2}{*}{\small{\FM${\langle{\tau_2^*}\rangle}{\rightarrow}${\M}$\langle${\TM}$\rangle$}}         & $\Delta$ Pr / $\Delta$ Re          & \textcolor{darkgreen}{\contour{darkgreen}{+}}1.5  /  \textcolor{red}{\contour{red}{-}}2.8  & \textcolor{darkgreen}{\contour{darkgreen}{+}}2.5  /  \textcolor{red}{\contour{red}{-}}6.6  & \textcolor{darkgreen}{\contour{darkgreen}{+}}2.1  /  \textcolor{red}{\contour{red}{-}}7.5  & \textcolor{darkgreen}{\contour{darkgreen}{+}}1.1  /  \textcolor{red}{\contour{red}{-}}8.7  & \textcolor{red}{\contour{red}{-}}4.1  /  \textcolor{red}{\contour{red}{-}}3.8 & 
    & \textcolor{darkgreen}{\contour{darkgreen}{+}}1.3  /  \textcolor{red}{\contour{red}{-}}1.7  & \textcolor{darkgreen}{\contour{darkgreen}{+}}2.7  /  \textcolor{red}{\contour{red}{-}}5.7  & \textcolor{darkgreen}{\contour{darkgreen}{+}}1.9  /  \textcolor{red}{\contour{red}{-}}6.2  & \textcolor{darkgreen}{\contour{darkgreen}{+}}2.3  /  \textcolor{red}{\contour{red}{-}}7.0  & \textcolor{darkgreen}{\contour{darkgreen}{+}}0.2  /  \textcolor{red}{\contour{red}{-}}3.9  \\ \cline{3-3}
     \cline{3-3}
                     &                 & $\%$ Filter / $\Delta$ F1  &	17.8 / \textcolor{red}{\contour{red}{-}}1.6 &	48.2 / \textcolor{red}{\contour{red}{-}} 6.5 &	59.8 / \textcolor{red}{\contour{red}{-}}8.9 &	75.7 / \textcolor{red}{\contour{red}{-}}12.2  &	91.7 / \textcolor{red}{\contour{red}{-}}7.0
  &    & 	15.2 / \textcolor{red}{\contour{red}{-}}0.8 &	48.7 / \textcolor{red}{\contour{red}{-}}5.4 &	57.3 / \textcolor{red}{\contour{red}{-}}7.2 &	72.0 / \textcolor{red}{\contour{red}{-}}9.5 &	93.8 / \textcolor{red}{\contour{red}{-}}7.1                    \\ \bottomrule

\end{tabular}
}
\vspace{-8pt}
\caption{Results showing effectiveness of question filtering. For each dataset and model {\M}, we train $6$ question filters: five \FMTval{i}'s for $i \in \{0.3,0.5,0.6,0.7,0.9\}$ and one {\FM}. For a particular filter {\F} operating with {\MTM}, $\Delta$ (Pr/Re/F1) refers to the difference in (Pr/Re/F1) of {\F}${\langle{\tau_2}^*\rangle}{\rightarrow}${\MTM} and {\MTM}. $\%$ Filter refers to the $\%$ of questions preemptively discarded by the question filter. {\MTM} results for AQAD are relative to {\M}$\langle 0 \rangle$.}
\label{tab:main_table}
\end{table*}
\begin{table*}[t]
\centering
\resizebox{0.7\linewidth}{!}{
\begin{tabular}{cccccccc}
\toprule
\multicolumn{3}{c}{\textbf{Threshold {\TM} for {\M} $\rightarrow$}}  & 0.3 & 0.5 & 0.6 & 0.7 & 0.9 \\ \midrule
\multirow{4}{*}{\begin{turn}{90}\textbf{WikiQA}\end{turn}} & \multirow{2}{*}{{\FW}} & RoBERTa-Base & 	0 / 0 &	0.2 / \textcolor{red}{\contour{red}{-}}0.1 &	0 / 0 & 0.2 / \textcolor{red}{\contour{red}{-}}0.1 &	0.8 / \textcolor{red}{\contour{red}{-}}0.8  \\ \cmidrule{3-3}
        
&& RoBERTa-Large & 	0 / 0 &	0 / 0 &	0.2 / \textcolor{red}{\contour{red}{-}}0.1 &	0.3 / \textcolor{red}{\contour{red}{-}}0.2 &	0 / 0  \\ \cmidrule{2-3}

& \multirow{2}{*}{{\FC}} & RoBERTa-Base &	4.1 / \textcolor{red}{\contour{red}{-}}1.2 &	1.5 / \textcolor{red}{\contour{red}{-}}0.8 &	2.1 / \textcolor{red}{\contour{red}{-}}0.8 &	4.1 / \textcolor{red}{\contour{red}{-}}0.9 &	15.6 / \textcolor{red}{\contour{red}{-}}1.7 \\  \cmidrule{3-3}
& & RoBERTa-Large & 	0.8 / \textcolor{red}{\contour{red}{-}}0.8 &	1.3 / \textcolor{red}{\contour{red}{-}}0.9 &	0.8 / \textcolor{red}{\contour{red}{-}}0.9 &	1.2 / \textcolor{red}{\contour{red}{-}}1.1 &	2.8 / \textcolor{red}{\contour{red}{-}}0.6 \\
\midrule
\multirow{4}{*}{\begin{turn}{90}\textbf{ASNQ}\end{turn}} & \multirow{2}{*}{{\FW}} & RoBERTa-Base & 	0.3 / \textcolor{red}{\contour{red}{-}}0.1 &	0.9 / \textcolor{red}{\contour{red}{-}}0.4 &	0.9 / \textcolor{red}{\contour{red}{-}}0.4 &	1.0 / \textcolor{red}{\contour{red}{-}}0.2 &	0 / 0  \\ \cmidrule{3-3}
        
&& RoBERTa-Large & 	0.1 / \textcolor{red}{\contour{red}{-}}0.2 &	0.1 / \textcolor{red}{\contour{red}{-}}0.1 &	0.1 / \textcolor{red}{\contour{red}{-}}0.2 &	0.2 / \textcolor{red}{\contour{red}{-}}0.1 &	0 / 0  \\ \cmidrule{2-3}

& \multirow{2}{*}{{\FC}} & RoBERTa-Base &	0.2 / \textcolor{red}{\contour{red}{-}}0.1 &	1.0 / \textcolor{red}{\contour{red}{-}}0.2 &	22.3 / \textcolor{red}{\contour{red}{-}}5.0 &	38.5 / \textcolor{red}{\contour{red}{-}}6.2 &	84.4 / \textcolor{red}{\contour{red}{-}}4.8 \\  \cmidrule{3-3}
& & RoBERTa-Large & 	0.3 / \textcolor{red}{\contour{red}{-}}0.2 &	0.5 / \textcolor{red}{\contour{red}{-}}0.3 &	2.7 / \textcolor{red}{\contour{red}{-}}0.8 &	7.2 / \textcolor{red}{\contour{red}{-}}1.3 &	62.5 / \textcolor{red}{\contour{red}{-}}8.1 \\ \midrule

\multirow{4}{*}{\begin{turn}{90}\textbf{SQuAD 1.1}\end{turn}} & \multirow{2}{*}{{\FW}} & RoBERTa-Base & 	0 / 0 &	0.4 / \textcolor{red}{\contour{red}{-}}0.1 &	0.5 / \textcolor{red}{\contour{red}{-}}0.2 &	0.7 / \textcolor{red}{\contour{red}{-}}0.3 &	0.2 / \textcolor{red}{\contour{red}{-}}0.2  \\ \cmidrule{3-3}
        
&& RoBERTa-Large & 	0.1 / \textcolor{red}{\contour{red}{-}}0.1 &	0.3 / \textcolor{red}{\contour{red}{-}}0.2 &	0.3 / \textcolor{red}{\contour{red}{-}}0.4 &	0.5 / \textcolor{red}{\contour{red}{-}}0.2 &	0.3 / \textcolor{red}{\contour{red}{-}}0.4  \\ \cmidrule{2-3}

& \multirow{2}{*}{{\FC}} & RoBERTa-Base &	0 / 0 &	0.3 / \textcolor{red}{\contour{red}{-}}0.1 &	28.0 / \textcolor{red}{\contour{red}{-}}9.1 &	34.3 / \textcolor{red}{\contour{red}{-}}8.7 &	59.7 / \textcolor{red}{\contour{red}{-}}9.4 \\  \cmidrule{3-3}
& & RoBERTa-Large & 	0 / 0 &	0.8 / \textcolor{red}{\contour{red}{-}}0.3 &	0.8 / \textcolor{red}{\contour{red}{-}}0.3 &	2.6 / \textcolor{red}{\contour{red}{-}}1.0 &	31.0 / \textcolor{red}{\contour{red}{-}}8.5 \\ \midrule

\multirow{4}{*}{\begin{turn}{90}\textbf{AQAD}\end{turn}} & \multirow{2}{*}{{\FW}} & RoBERTa-Base & 	0 / 0 &	4.8 / \textcolor{red}{\contour{red}{-}}0.7 &	4.2 / \textcolor{red}{\contour{red}{-}}0.4 &	15.7 / \textcolor{red}{\contour{red}{-}}1.9 &	32.6 / \textcolor{red}{\contour{red}{-}}2.5  \\ \cmidrule{3-3}
        
&& RoBERTa-Large & 	0 / 0 &	4.3 / \textcolor{red}{\contour{red}{-}}1.1 &	4.4 / \textcolor{red}{\contour{red}{-}}0.8 &	15.6 / \textcolor{red}{\contour{red}{-}}2.3 &	29.6 / \textcolor{red}{\contour{red}{-}}3.2  \\ \cmidrule{2-3}

& \multirow{2}{*}{{\FC}} & RoBERTa-Base &	4.5 / \textcolor{red}{\contour{red}{-}}0.8 &	18.5 / \textcolor{red}{\contour{red}{-}}3.1 &	38.5 / \textcolor{red}{\contour{red}{-}}5.8 &	38.9 / \textcolor{red}{\contour{red}{-}}4.5 &	82.3 / \textcolor{red}{\contour{red}{-}}3.6 \\  \cmidrule{3-3}
& & RoBERTa-Large & 	3.1 / \textcolor{red}{\contour{red}{-}}0.4 &	20.2 / \textcolor{red}{\contour{red}{-}}3.3 &	36.1 / \textcolor{red}{\contour{red}{-}}5.4 &	37.2 / \textcolor{red}{\contour{red}{-}}4.3 &	80.9 / \textcolor{red}{\contour{red}{-}}3.4 \\ 
\bottomrule
\end{tabular}
}
\vspace{-8pt}
\caption{Table presenting performance of baseline question filters {\FW} and {\FC} on all four datasets corresponding to their optimal operating threshold {\TFBest}.} 
\label{tab:app_baselines}
\end{table*}

\vfill

\end{document}